\documentclass[9.5pt,journal,compsoc]{IEEEtran}

\ifCLASSOPTIONcompsoc
  \usepackage[nocompress]{cite}
\else
  \usepackage{cite}
\fi
\ifCLASSINFOpdf
\else
\fi

\usepackage{times}
\usepackage{epsfig}
\usepackage{amsmath}
\usepackage{multirow}
\usepackage{hhline}
\usepackage{caption}
\usepackage{rotating}
\usepackage{enumitem }
\usepackage{color}
\usepackage{chngpage}
\usepackage{hyperref}
\usepackage{threeparttable}

\usepackage{xcolor}

\newcommand{\colorchunk}[1]{#1}

\usepackage{makecell}

\usepackage[switch]{lineno} 

\usepackage{amsmath,amssymb,amsopn}
\usepackage{bm} 
\usepackage{mathtools}

\newcommand*\rot{\rotatebox{90}}

\usepackage{pifont}
\newcommand{\cmark}{\ding{51}}%
\newcommand{\xmark}{\text{\ding{53}}}%

\begin{document}
%
\title{A Curriculum Domain Adaptation Approach to the Semantic Segmentation of Urban Scenes}
%
%
%
%

\author{Yang~Zhang,~\IEEEmembership{Student Member,~IEEE,}
        Philip~David,~\IEEEmembership{Member,~IEEE,}
        Hassan Foroosh,~\IEEEmembership{Senior Member,~IEEE,}
        and~Boqing~Gong,~\IEEEmembership{Member,~IEEE}
}

\IEEEtitleabstractindextext{%
\begin{abstract}
During the last half decade, convolutional neural networks (CNNs) have triumphed over semantic segmentation, which is one of the core tasks in many applications such as autonomous driving and augmented reality. However, to train CNNs requires a considerable amount of data, which is difficult to collect and laborious to annotate. Recent advances in computer graphics make it possible to train CNNs on photo-realistic synthetic imagery with computer-generated annotations. Despite this, the domain mismatch between the real images and the synthetic data hinders the models' performance. Hence, we propose a curriculum-style learning approach to minimizing the domain gap in urban scene semantic segmentation. The curriculum domain adaptation solves easy tasks first to infer necessary properties about the target domain; in particular, the first task is to learn global label distributions over images and local distributions over landmark superpixels. These are easy to estimate because images of urban scenes have strong idiosyncrasies (e.g., the size and spatial relations of buildings, streets, cars, etc.). We then train a segmentation network, while regularizing its predictions in the target domain to follow those inferred properties.  In experiments, our method outperforms the baselines on two datasets and two backbone networks. We also report extensive ablation studies about our approach.


\end{abstract}

\begin{IEEEkeywords}
Domain Adaptation, Semantic Segmentation, Curriculum Learning, Deep Learning, Self-Driving
\end{IEEEkeywords}}

\maketitle

\IEEEdisplaynontitleabstractindextext

%
\IEEEpeerreviewmaketitle

\let\thefootnote\relax\footnote{Code is available at \url{https://github.com/YangZhang4065/AdaptationSeg}.}


\IEEEraisesectionheading{\section{Introduction}\label{sec:introduction}}
\IEEEPARstart{S}{emantic} segmentation is one of the most challenging and fundamental problems in computer vision. It assigns a semantic label to each pixel of an input image~\cite{garcia2017review}. The resulting output is a dense and rich annotation of the image, with one semantic label per pixel. Semantic segmentation  facilitates many downstream applications, including autonomous driving,  which, over the past few years, has made great strides towards the use by the general population. Indeed, several datasets and test suites have been developed for research on autonomous driving~\cite{cordts2016cityscapes,xu2016end,airsim2017fsr,Dosovitskiy17,richter_playing_2016} and, among them, semantic segmentation is often considered one of the key tasks. 

Convolutional neural networks (CNNs)~\cite{long_fully_2015,KrizhevskyNIPS12Imagenet} have become a hallmark backbone model to solve the semantic segmentation of large-scale image sets over the last half decade. All of the top-performing methods on the challenge board of the Cityscapes pixel-level semantic labeling task~\cite{cordts2016cityscapes} rely on CNNs. One of the reasons that CNNs are able to achieve a high level of accuracy for this task is that the training set is sufficiently large and well-labeled, covering the variability of the test set for the research purpose. In practice, however, it is often hard to acquire new training sets that fully cover the huge variability of real-life test scenarios. Even if one could compose a large-scale dataset with sufficient variability, it would be extremely tedious to label the images with pixel-wise semantic labels. For example, Cordts et al.\ report that the annotation and quality control took more than 1.5 hours per image in the popular Cityscapes dataset~\cite{cordts2016cityscapes}.

These challenges motivate researchers to approach the segmentation problem by using complementary synthetic data. With modern graphics engines, automatically synthesizing diverse urban-scene images along with pixel-wise labels would require very little to zero human labor. Figure~\ref{fGTAQualitative} shows some synthetic images of the GTA dataset~\cite{richter_playing_2016}. They are quite photo-realistic, giving rise to the hope that a semantic segmentation neural network trained from them can perform reasonably well on the real images as well. However, our experiments show that this is not the case (cf.\ Section~\ref{sExperiments}), signifying a severe mismatch between the real images and the synthesized ones. Multiple factors may contribute to the mismatch, such as the scene layout, capture device (camera vs.\ rendering engine), view angles, lighting conditions and shadows, textures, etc.  


In this paper, our main objective is to investigate the use of domain adaptation techniques~\cite{hoffman2017cycada,hoffman_fcns_2016,zhang2017curriculum,peng2017visda} to more effectively transfer the semantic segmentation neural networks trained using synthetic images to high-quality segmentation networks for real images. We build our efforts upon our prior work~\cite{zhang2017curriculum}, in which we propose a novel domain adaptation approach to the semantic segmentation of urban scenes. 

Domain adaption, which mainly aims to boost models' performance when the target domain of interest  differs from the one where the models are trained, has long been a popular topic in machine learning and computer vision~\cite{csurka2017domain}. It has recently drawn even greater attention along with transfer learning thanks to the prevalence of deep neural networks which are often ``data-hungry''.  An intuitive domain adaptation strategy is to learn domain-invariant feature representations for the images of both domains, where the source domain supplies a labeled training set and the target domain reveals zero to a few labeled images along with many unlabeled ones. In this case, the source domain features would resemble the target ones' characteristics. Thus, the model trained on the labeled source domain can be generalized to the target domain. Earlier ``shallow'' methods achieve such goals by exploiting various intrinsic structures of the data~\cite{gong2012geodesic,GopalanICCV11Domain,FernandoICCV13Unsupervised,SunAAAI16Return,PanTNN11Domain,GongICML13Connecting,AljundiCVPR15Landmarks,KulisCVPR11What}. In contrast, the recent ``deep'' methods mainly devise new loss functions and/or network architectures to add domain-invariant ingredients to the gradients backpropagating through the neural networks~\cite{TzengX14Deep,LongICML15Learning,tzeng_simultaneous_2015,GaninX15Domainadversarial,GaninICML15Unsupervised}.

Upon observing the success of learning domain-invariant features in the prior domain adaptation tasks, it is a natural tendency to follow the same principle for the adaptation of semantic segmentation models. There have been some positive results along this line~\cite{hoffman_fcns_2016,peng2017visda}. However, the underlying assumption of this principle may prevent the methods designed around it from achieving high adaptation performance. By focusing on learning domain-invariant features $X$ (i.e., such that $P_S(X)\approx P_T(X)$, where the subscripts $S$ and $T$ stand for the source and target domains, respectively), one assumes  the conditional distribution $P(Y|X)$, where $Y$ are the pixel labels, is more or less shared by the two domains. This assumption is less likely to be true when the classification boundary becomes more and more sophisticated --- the prediction function for semantic segmentation has to be sophisticated. The sets of pixel labels are high-dimensional, highly structured, and interdependent, implying that the learner has to resolve the predictions in an exponentially large label space.  Besides, some discriminative cues in the data would be suppressed if one matches the feature representations of the two domains without taking careful account of the structured labels. Finally, data instances are the proxy to measure the domain difference~\cite{GrettonBC09Covariate,GaninICML15Unsupervised,GaninX15Domainadversarial}. However, it is not immediately clear what comprises the instance in semantic segmentation~\cite{hoffman_fcns_2016}, especially given that the top-performing segmentation methods  are built upon deep neural networks~\cite{long_fully_2015,pathak_constrained_2015,noh_learning_2015,chen_semantic_2014}. Hoffman et al.\ take each spatial unit in the fully convolutional network (FCN)~\cite{long_fully_2015} as an instance~\cite{hoffman_fcns_2016}. We contend that such instances are actually non-i.i.d.\ in either individual  domain, as their receptive fields overlap with each other. 

How can we avoid the assumption that the source and target domains share the same prediction function in a transformed domain-invariant feature space? Our proposed solution draws on two key observations. One is that the urban traffic scene images have strong idiosyncrasies (e.g., the size and spatial relations of buildings, streets, cars, etc.). Therefore, \emph{some tasks are ``easy'' and, more importantly, suffer less because of the domain discrepancy}. For instance, it is easy to infer from a traffic scene image that the road often occupies a larger number of pixels than the traffic sign does. Second, the structured output in semantic segmentation enables convenient posterior regularization~\cite{ganchev2010posterior}, as opposed to the generic (e.g., $\ell_2$) regularization over model parameters. 

Accordingly, we propose a curriculum-style~\cite{bengio2009curriculum} domain adaptation approach. Recall that, in domain adaptation, only the source domain supplies many labeled data while there are no or only scarce labels from the target domain. Our curriculum domain adaptation begins with the easy tasks, in order to gain some high-level properties about the unknown pixel-level labels for each target image. It then learns a semantic segmentation network, the hard task, whose predictions over the target images are constrained to follow those target-domain properties as much as possible. 

To develop the easy tasks for the curriculum, we consider estimating label distributions over both global images and some landmark superpixels of the target domain. Take the former for instance. The label distribution indicates the percentage of pixels in an image that correspond to each category. We argue that these tasks are easier, despite the domain mismatch, than predicting pixel-wise labels. The label distributions are only rough estimations about the labels's statistics. Moreover, the size relations between road, building, sky, people, etc.\ constrain the shape of the distributions, effectively reducing the search space. Finally, models to estimate the label distributions over superpixels may benefit from the urban scenes' canonical layout that transcends domains, e.g., buildings stand beside streets. 

Why and when are these seemingly simple label distributions useful for the domain adaptation of semantic segmentation? In our experiments, we find that the segmentation networks trained on the source domain perform poorly on many target images, giving rise to disproportionate label assignments (e.g., many more pixels are classified to sidewalks than to streets). To rectify this, the image-level label distribution informs the segmentation network \emph{how} to update the predictions while the label distributions of the anchor superpixels tell the network \emph{where} to update. Jointly, they guide the adaptation of the networks to the target domain to, at least, generate proportional label predictions. Note that additional ``easy tasks'' can be incorporated into our approach in the future. 

Our main contribution is the proposed curriculum-style domain adaptation for the semantic segmentation of urban scenes. We select for the curriculum the easy and useful tasks of inferring label distributions for both target images and landmark superpixels in order to gain some necessary properties about the target domain. Built upon these, we learn a pixel-wise discriminative segmentation network from the labeled source data and, meanwhile, conduct a ``sanity check'' to ensure the network behavior is consistent with the previously learned knowledge about the target domain. Our approach effectively eludes the assumption about the existence of a common prediction function for both domains in a transformed feature space. It readily applies to different segmentation networks as it does not change the network architecture or impact any intermediate layers. 

Beyond our prior work~\cite{zhang2017curriculum}, we provide more algorithmic details and experimental studies about our approach, including new experiments using the GTA dataset~\cite{richter_playing_2016} and ablation studies about the number of superpixels, feature representations of the superpixels, various backbone neural networks, prediction confusion matrix, etc. In addition, we introduce a color constancy scheme into our framework, which significantly improves the adaptation performance and may be plugged into any domain adaptation method as a standalone image pre-processing step. We also quantitatively measure the ``market value'' of the synthetic data to reveal how much cost it could save out of the labeling of real images. Finally, we provide a comprehensive survey about the works published after ours~\cite{zhang2017curriculum} on the domain adaptation for semantic segmentation. We group them into different categories and experimentally demonstrate that other methods are complementary to ours.

\section{Related work}
We broadly discuss related work on domain adaptation and semantic segmentation in this section. Section~\ref{SecReview} provides a more focused review about the domain adaptation methods for semantic segmentation, along with experimental studies about the complementary effect between them and ours of different categories.

\subsection{Domain adaptation} Conventional machine learning algorithms rely on the standard assumption that the training and test data are drawn i.i.d.\ from the same underlying distribution. However, it is often the case that there exists some discrepancy between the training and test stages. Domain adaptation aims to rectify this mismatch and tune the models toward better generalization at the test stage~\cite{TorralbaCVPR11Unbiased,TommasiX15Deeper,GongLSVRR12Overcoming,KhoslaECCV12Undoing,GrettonBC09Covariate}.


The existing work on domain adaptation mostly focuses on classification and regression problems~\cite{PatelSPM15Visual,PanTKDE10Survey}, e.g., learning from online images to classify real world objects~\cite{saenko2010adapting,gong2012geodesic}, and, more recently, aims to improve the adaptability of deep neural networks~\cite{LongICML15Learning,GaninX15Domainadversarial,GaninICML15Unsupervised,tzeng_simultaneous_2015,bousmalis2016unsupervised}. Among them, the most relevant works to ours are those exploring simulated data~\cite{SunBMVC14Virtual,XuX14Hierarchical,ros_synthia_2016,VazquezPAMI14Virtual,hoffman_fcns_2016,peng2017synthetic,shrivastava2016learning}. Sun and Saenko train generic object detectors from synthetic images~\cite{SunBMVC14Virtual}, while Vazquez et al.\ use virtual images to improve pedestrian detections in real environments~\cite{VazquezPAMI14Virtual}. The other way around, i.e., how to improve the quality of the simulated images using the real ones, is studied in~\cite{shrivastava2016learning,peng2017synthetic}.


\subsection{Semantic segmentation}
Semantic segmentation is the task of assigning an object label to each pixel of an image. Traditional methods~\cite{shotton_semantic_2008,tighe2010superparsing,zhang2010semantic} rely on local image features manually designed by domain experts. After the pioneering works~\cite{chen_semantic_2014,long_fully_2015} that introduces the convolutional neural network (CNN)~\cite{Lecun98Gradient} to semantic segmentation, most recent top-performing methods are also built on CNNs~\cite{wu_wider_2016,ros_training_2016,badrinarayanan_segnet:_2015,zhao_pyramid_2016,noh_learning_2015,dai_instance-aware_2016}. 

Currently, there are multiple and increasing numbers of semantic segmentation datasets aiming for different computer vision applications. Some general ones include the PASCAL VOC2012 Challenge~\cite{everingham_pascal_2015}, which contains nearly 10,000 annotated images for the segmentation competition, and the MS COCO Challenge~\cite{lin_microsoft_2014}, which includes over 200,000 annotated images. In our paper, we focus on urban outdoor scenes. Several urban scene segmentation datasets are publicly available such as Cityscapes~\cite{cordts2016cityscapes}, a vehicle-centric dataset created primarily in German cities, KITTI~\cite{geiger2013vision}, another vehicle-centric dataset captured in the German city Karlsruhe, Berkeley DeepDrive Video Dataset~\cite{xu2016end}, a dashcam dataset collected in United States,  Mapillary Vistas Dataset~\cite{neuhold2017mapillary}, so far known as the largest outdoor urban scene segmentation dataset collected from all over the world, WildDash, a much smaller yet diverse dataset for benchmark purpose, and CamVid~\cite{BrostowSFC:ECCV08}, a small and low-resolution toy dashcam dataset. An enormous amount of labor-intensive work is required to annotate the  images that are needed to obtain accurate segmentation models. According to \cite{richter_playing_2016}, it took about 60 minutes to manually segment each image in \cite{brostow_semantic_2009} and about 90 minutes for each in ~\cite{cordts2016cityscapes}. A plausible approach to reducing the human annotation workload is to utilize weakly supervised information such as image labels and bounding boxes~\cite{pathak_constrained_2015,hong_learning_2016,PapandreouICCV15Weakly,pinheiro_image-level_2015}. 

We instead explore the almost labor-freely labeled virtual images for training high-quality segmentation networks. In \cite{richter_playing_2016}, annotating a synthetic image took only 7 seconds on average through a computer game. For the urban scenes, we use the SYNTHIA~\cite{ros_synthia_2016} and GTA~\cite{richter_playing_2016} datasets which contain images of virtual cities. Although not used in our experiments, another synthetic segmentation dataset worth mentioning is Virtual KITTI~\cite{Gaidon:Virtual:CVPR2016}, a synthetic  duplication of the original  KITTI~\cite{geiger2013vision}  dataset.

\subsection{Domain adaptation for semantic segmentation} Due to the clear visual mismatch between synthetic and real data~\cite{shafaei_play_2016,richter_playing_2016,ros_synthia_2016}, we expect to use domain adaptation to enhance the segmentation performance on real images by networks trained on synthetic imagery. To the best of our knowledge, our work~\cite{zhang2017curriculum} and the FCNs in the wild~\cite{hoffman_fcns_2016} are among the very first attempts to tackle this problem. It subsequently became an independent track in the Visual Domain Adaptation Challenge (VisDA) 2017~\cite{peng2017visda}. After that, some feature-alignment based methods are developed~\cite{sankaranarayanan2017unsupervised,hoffman2017cycada,saito2017maximum,chen2017road,wu2018dcan}. We postpone further discussion to Section~\ref{SecReview}, which presents a comprehensive survey about the works published after ours and before December 2018. We group them into two major categories and describe their main methods. We also summarize their results by Table~\ref{Tresults_SOTA} and analyze their complementary relationship with ours.

\section{Approach} \label{sApproach}

In this section, we present the details of our approach to curriculum domain adaptation for the semantic segmentation of urban scene images. Unlike previous works~\cite{PatelSPM15Visual,hoffman_fcns_2016} that align the domains via an intermediate feature space and thereby implicitly assume the existence of a single decision function for the two domains, it is our intuition that, for structured prediction (i.e., semantic segmentation here), the cross-domain generalization of machine learning models can be more efficiently improved if we avoid this assumption and instead train them subject to necessary properties they should retain in the target domain. After a brief introduction on the preliminaries, we will present how to facilitate semantic segmentation adaptation during training using estimated target domain properties in  Section~\ref{sSSMwTP}. Then we will focus on the types of target domain properties and how to estimate them in  Section~\ref{sTargetProperty}.

\subsection{Preliminaries}
In particular, the properties of interest concern pixel-wise category labels $Y_t\in\mathbb{R}^{W\times H \times C}$ of an arbitrary image $I_t\in\mathbb{R}^{W\times H}$ from the target domain, where $W$ and $H$ are the width and height of the image, respectively, and $C$ is the number of categories. We use one-hot vector encoding for the groundtruth labels, i.e., $Y_t(i,j,c)$ takes the value of either 0 or 1, where the latter means that the $c$-th label is assigned by a human annotator to the pixel at $(i,j)$. Correspondingly, the prediction $\widehat{Y}_t(i,j,c)\in[0,1]$ by a segmentation network is realized by a softmax function per pixel. 


We express each target property in the form of a distribution $p_t$ over the $C$ categories, where $\sum^C_c p_t(c)=1$ and $0\leqslant p_t(c), \forall c$. $p_t(c)$ represents the occupancy proportion of the category $c$ over the $t$-th target image or a superpixel of that image. Therefore, one can immediately calculate the distribution $p_t$ given the human annotations $Y_t$ to the image. For instance, the image-level label distribution is expressed by
\begin{align}
    p_t(c) = \frac{1}{WH}\sum_{i=1}^W\sum_{j=1}^H Y_t(i,j,c), \quad \forall c. \label{eP}
\end{align}
Similarly, we can compute the estimated target property/distribution from the network predictions $\widehat{Y}_t$ and denote it by $\widehat{p}_t$,
\begin{align}
    \widehat{p}_t(c)=   \frac{1}{WH}\sum_{i=1}^W\sum_{j=1}^H \left(\dfrac{\widehat{Y}_t(i,j,c)}{\max_{c'}(\widehat{Y}_t(i,j,c'))}\right)^K,\quad \forall c \label{eConv}
\end{align}
where $K>1$ is a large constant whose effect is to ``sharpen'' the softmax activation per pixel $\widehat{Y}(i,j,c)$ such that the summand is either 1 or very close to 0, in a similar shape as the summad $Y_t(i,j,c)$ of eq.~(\ref{eP}). We set $K=6$ in our experiment as larger $K$ caused numerical instability. Finally, we $\ell_1$-normalize the vector $(\widehat{p}_t(1),\widehat{p}_t(2),\cdots,\widehat{p}_t(C))^T$ such that its elements are all greater than 0 and sum up to 1 --- in other words, the vector remains a valid distribution.

\subsection{Domain adaptation observing the target properties}\label{sSSMwTP}
Ideally, we would like to have a segmentation network to imitate human annotators of the target domain. Therefore, necessarily, the properties of their annotation results should be the same too. We capture this notion by minimizing the cross entropy
$\mathcal{C}(p_t,\widehat{p}_t)=H(p_t)+\textsc{KL}(p_t,\widehat{p}_t)$ at training, where the first term of the right-hand side is the entropy and the second is the KL-divergence. 

Given a mini-batch consisting of both source images ($S$) and target images ($T$), the overall objective function for training the cross-domain generalizing segmentation network is
\begin{align}
    \min \; \dfrac{\gamma}{|S|}\sum_{s\in S}\mathcal{L}\Big(Y_s,\widehat{Y}_s\Big) + \frac{1-\gamma}{|T|}\sum_{t\in T}\sum_{k}\mathcal{C}\Big(p^k_t,\widehat{p}^k_t\Big)  \label{eLoss}
\end{align}
where $\mathcal{L}$ is the pixel-wise cross-entropy loss defined over the fully labeled source domain images, enforcing the network to have the pixel-level discriminative capabilities, and the second term is over the unlabeled target domain images, hinting the network what necessary properties its predictions should have in the target domain. We use  $\gamma\in[0,1]$ to balance the two strengths in training and superscript $k$ to index different types of label distributions  (cf.\ $p_t$ in eq.~(\ref{eP}) and Section~\ref{sTargetProperty}).

Note that, in the unsupervised domain adaptation context, we actually cannot directly compute the label distributions $\{p^k_t\}$ because the  groundtruth annotations of the target domain are unknown. Nonetheless, using the labeled source domain data, these distributions are easier to estimate than are the labels for every pixel of a target image. We present two types of such properties and the details for inferring them  in the next section. In future work, it is worth exploring other properties.


\textbf{Remarks.} 
Mathematically, the objective function has a similar form as  model compression~\cite{bucilua2006model,hinton2015distilling}. Hence, we borrow some concepts to gain a more intuitive understanding of our domain adaptation procedure. The ``student'' network follows a curriculum to learn simple knowledge about the target domain before it addresses the hard one of semantically segmenting images. The models inferring the target properties act like ``teachers'', as they hint what label distributions the final solution (image annotation) may have in the target domain at the image and superpixel levels. 

Another perspective is to understand the target properties as a posterior regularization~\cite{ganchev2010posterior} for the network. The posterior regularization can conveniently encode a priori knowledge into the objective function. Some applications using this approach include weakly supervised segmentation~\cite{pathak_constrained_2015,ros_training_2016} and detection~\cite{BilenBMVC14weakly} and rule-regularized training of neural networks~\cite{hu2016harnessing}. In addition to the domain adaptation setting and novel target properties,
another key distinction of our work is that we decouple the label distributions from the network predictions and thus avoid the EM type of optimization, which is often involved and incurs extra computational overhead. Our approach learns the segmentation network with almost effortless changes to the popular deep learning tools.

\subsection{Inferring the target properties} \label{sTargetProperty}
Thus far, we have presented the ``hard'' task --- learning the segmentation neural network --- in the curriculum domain adaptation. In this section, we describe the ``easy'' tasks, i.e., how to infer the target domain properties  without any annotations from the target domain. Our contributions also include selecting the particular form of label distributions to constitute the simple tasks.

\subsubsection{Global label distributions over images} \label{subsLD}
Due to the domain disparity, a baseline segmentation network trained on the source domain (i.e., using the first term of eq.~(\ref{eLoss})) could be easily crippled given the target images. In our experiments, we find that our baseline network constantly mistakes streets for sidewalks and/or cars (cf.\ Figure~\ref{fQualitative}). Consequently, the predicted labels for the pixels are highly disproportionate. 

To rectify this, we employ the label distribution $p_t$ over the global image as our first property (cf.\ eq.~(\ref{eP})). Without access to the target labels, we have to train machine learning models from the labeled source images to estimate the label distribution $p_t$ of a target image. Nonetheless, we argue that this is less challenging than generating  per-pixel predictions despite that both tasks are influenced by the domain mismatch.


In our experiments, we examine several different  approaches to this task. We extract 1536D image features from the output of the  average pooling layer in {Inception-Resnet-v2}~\cite{szegedy_inception-v4_2016} as the input to the following models.
\begin{description}
\item[Logistic regression.] Although multinomial logistic regression (LR) is mainly used for classification, its output is actually a valid distribution over the categories. For our purpose, we thus train it by replacing the one-hot vectors in the cross-entropy loss with the groundtruth label distribution $p_s$, which is counted by using eq.~(\ref{eP}) from the human labels of the source domain. Given a target image, we directly take the LR's output as the estimated label distribution $p_t$.

\item[Mean of nearest neighbors.]
We also test a nonparametric method by simply retrieving multiple nearest neighbor (NN) source images for each target image and then transferring the mean of their label distributions to the target image. We use the $\ell_2$ distance in the Inception-Resnet-v2 feature space for the NN retrieval.

\end{description}

Finally, we include two dumb predictions as the control experiments. One is, for any target image, to output the mean of all the label distributions in the source domain (\textbf{source mean}), and the other is to output a \textbf{uniform distribution}.

\subsubsection{Local label distributions of landmark superpixels} \label{subsSP}
The image-level label distribution globally penalizes potentially disproportional segmentation output in the target domain. It is yet inadequate in providing spatial regularization to the network. In this section, we consider the use of label distributions over some superpixels as the anchors to drive the network towards spatially desired target properties. 

Note that it is not necessary, and is even harmful, to use all of the superpixels in a target image to regularize the segmentation network because it would be too strong a force and might overrule the pixel-wise discriminativeness (obtained from the fully labeled source domain), especially when the label distributions are not inferred accurately enough. 

In order to have the dual effect of both estimating the label distributions of some superpixels and selecting them from all candidate superpixels, we  employ a linear SVM in this work. We first segment each image into 100 superpixels using linear spectral clustering~\cite{li_superpixel_2015}. For the superpixels of the source domain, we are able to assign a single dominant label to each of them and then train a multi-class SVM using the ``labeled'' superpixels of the source domain. Given a test superpixel of a target image, the multi-class SVM returns a class label as well as a decision value, which is interpreted as the confidence score about classifying this superpixel. We keep the top 30\% most confident superpixels in the target domain. The class labels are then encoded into one-hot vectors which serve as valid distributions about the category labels upon the selected landmark superpixels' area. Albeit simple, we find this method works very well.

In order to train the aforementioned superpixel SVM, we need to find a  way to represent the superpixels in a feature space. We encode both visual and contextual information to represent a superpixel. First, we use the FCN-8s~\cite{long_fully_2015} pre-trained on the PASCAL CONTEXT \cite{mottaghi_role_2014} dataset, which has 59 distinct classes, to obtain 59 detection scores for each pixel. We then average these scores within each superpixel. The final feature representation of a superpixel is a 295D concatenation of the 59D vectors of itself, its left and right superpixels, as well as the two respectively above and below it. As this feature representation relies on extra data source, we also examine handcrafted features and VGG features~\cite{simonyan_very_2014} in the experiments.

\begin{figure}
\centering
    \includegraphics[width=0.45\textwidth]{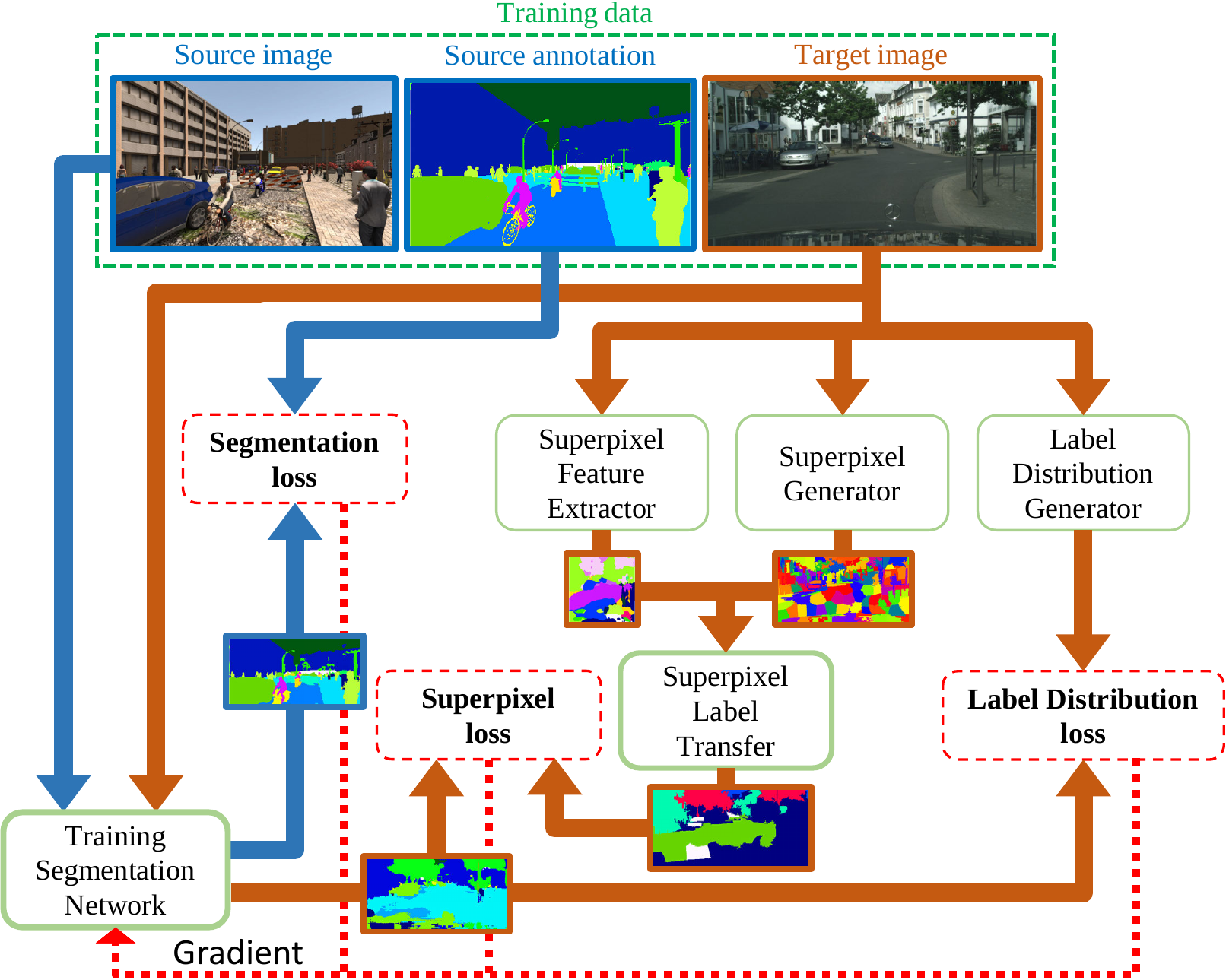}
  \caption{The overall framework of our curriculum domain adaptation approach to the semantic segmentation of urban scenes.} \label{fOverview}
\end{figure}

\subsection{Curriculum domain adaptation: recapitulation} \label{sRecap}
We recap the proposed curriculum domain adaptation using Figure~\ref{fOverview} before presenting the experiments in the next section. Our main idea is to execute the domain adaptation step by step, starting from the easy tasks that, compared to semantic segmentation, are less sensitive to domain discrepancy. We choose the label distributions over global images and local landmark superpixels in this work; more tasks will be explored in the future. The solutions to them provide useful gradients originating from the target domain (cf.\ the arrows with brown color in Figure~\ref{fOverview}), while the  source domain feeds the network with well-labeled images and segmentation masks (cf.\ the dark blue arrows in Figure~\ref{fOverview}).

\subsection{Color Constancy} \label{subsCC}

\begin{figure}
\centering
    \includegraphics[width=0.45\textwidth]{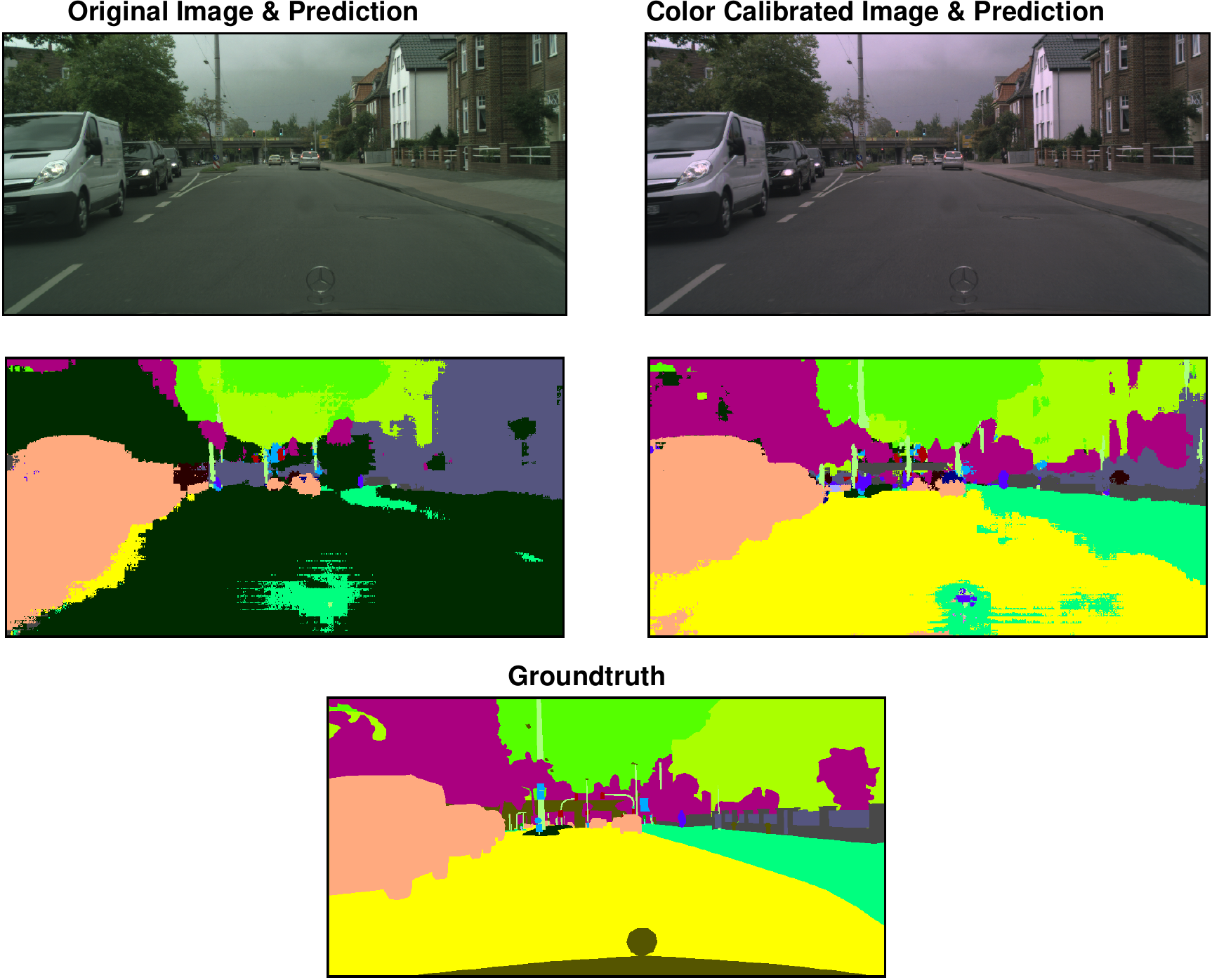}
  \caption{Predictions by the same  FCN-8s model, without domain adaptation, before and after calibrating the image's colors.}
  \label{fCC}
\end{figure}

In this subsection, we propose an color calibration preprocessing step which we find very effective in adapting semantic segmentation methods from the source synthetic domain to the target real image domain. We assume the colors of the two domains are drawn from different distributions. We calibrate the target domain images' colors to those of the source domain, hence reducing their discrepancy in terms of the color. We describe it here as an independent subsection because it can stand alone and can be added to any existing methods for the domain adaptation of semantic segmentation.

Humans have the ability to perceive the same color of an object even when it is exposed in different illuminations~\cite{foster2011color}, but image capturing sensors do not. As a result, different illuminations result in distinct RGB images captured by the cameras. Consequently, the perception incoherence hinders the performance of computer vision algorithms~\cite{barata2015improving} because the illumination is often among the key factors that cause the domain mismatch. To this end, we propose to use computational color constancy~\cite{gijsenij2011computational} to eliminate the influence of illuminations.

The goal of color constancy is to correct the colors of images acquired under unconventional or biased lighting sources to the colors that are supposed to be under the reference lighting condition. In our domain adaptation scenario, we assume that the source domain resembles the reference lighting condition. We learn a parametric model to describe both the target and source lighting sources and then try to restore the target images according to the source domain's light. However, not all the color constancy methods are applicable. For instance, some methods rely on physics priors or the statistics of natural images, both of which are unavailable in synthetic images. 

Due to the above concerns, we instead use a gamut-based color constancy method~\cite{gijsenij2010generalized} to align the target and source images in terms of their colors. This method infers the property of the light source under the assumption  that only a limited range of color could be observed under a certain light source. This matches our assumption that the target images' colors and the source images' colors belong to different distributions/ranges. In addition to the pixel values, the image edge and derivatives are also used to find the mapping.  While we omit the details of this color constancy method and refer the readers to~\cite{gijsenij2010generalized} instead, we show in Figure~\ref{fCC} how sensitive the segmentation model is to the illumination. We can see that, prior to applying color constancy, a large part of a CityScapes image is incorrectly classified as ``terrain'' only because the image is a little greenish.

\begin{sidewaysfigure*}
\small
\centering
    \includegraphics[height=0.75\textheight,width=1.0\textwidth]{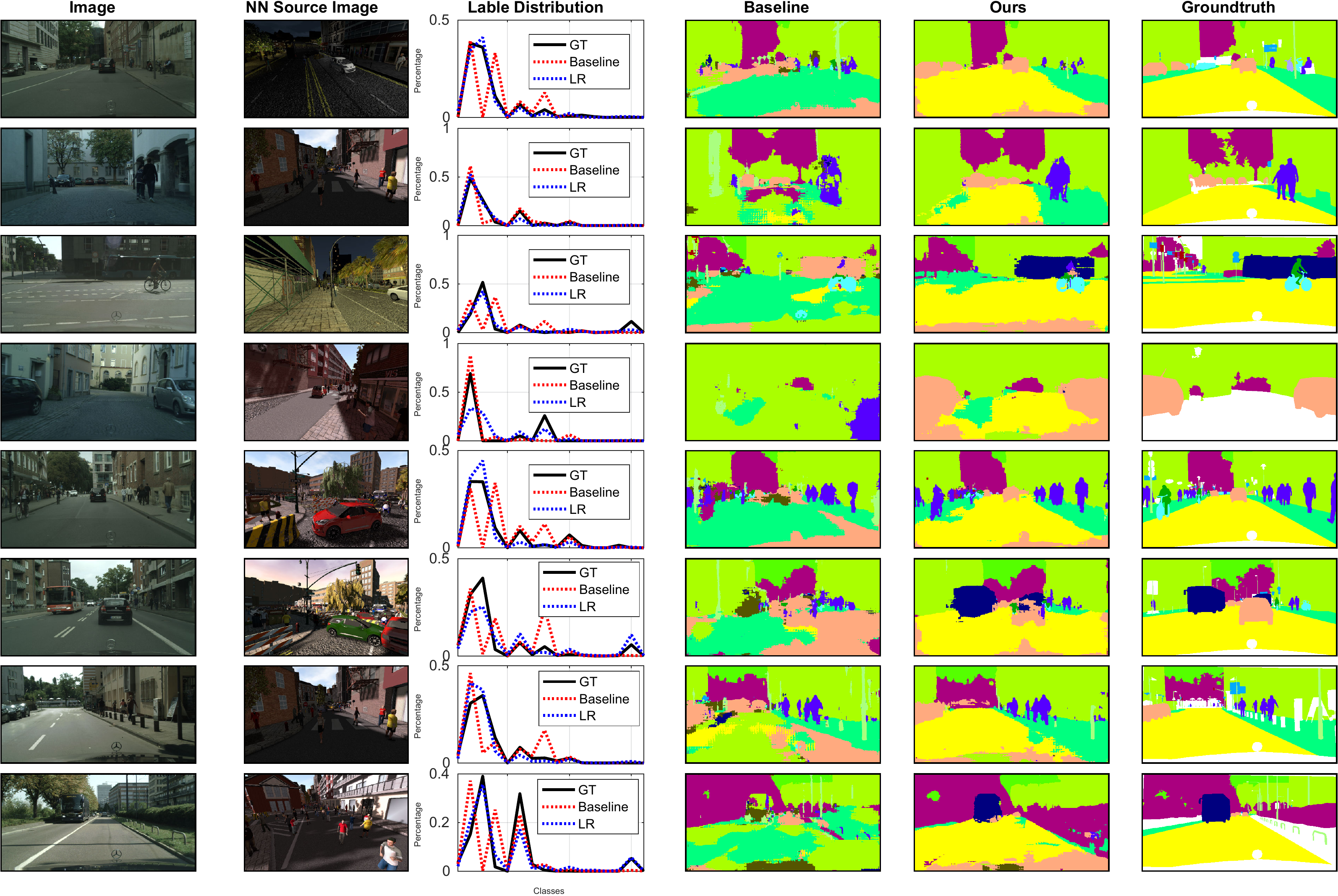}
    \vspace{-20pt}
  \caption{Qualitative semantic segmentation results on the Cityscapes dataset~\cite{ros_synthia_2016} (target domain). For each target image in the first column, we retrieve its nearest neighbor from the SYNTHIA~\cite{cordts2016cityscapes} dataset (source domain). The third column plots the label distributions due to the groundtruth pixel-wise semantic annotation, the predictions by the baseline network with no adaptation, and the inferred distribution by logistic regression. The last three columns are the segmentation results by the baseline network, our domain adaptation approach, and human annotators, respectively.} \label{fQualitative}
  \vspace{-15pt}
\end{sidewaysfigure*}

\begin{sidewaysfigure*}
\small
\centering
    \includegraphics[height=0.75\textheight,width=1.0\textwidth]{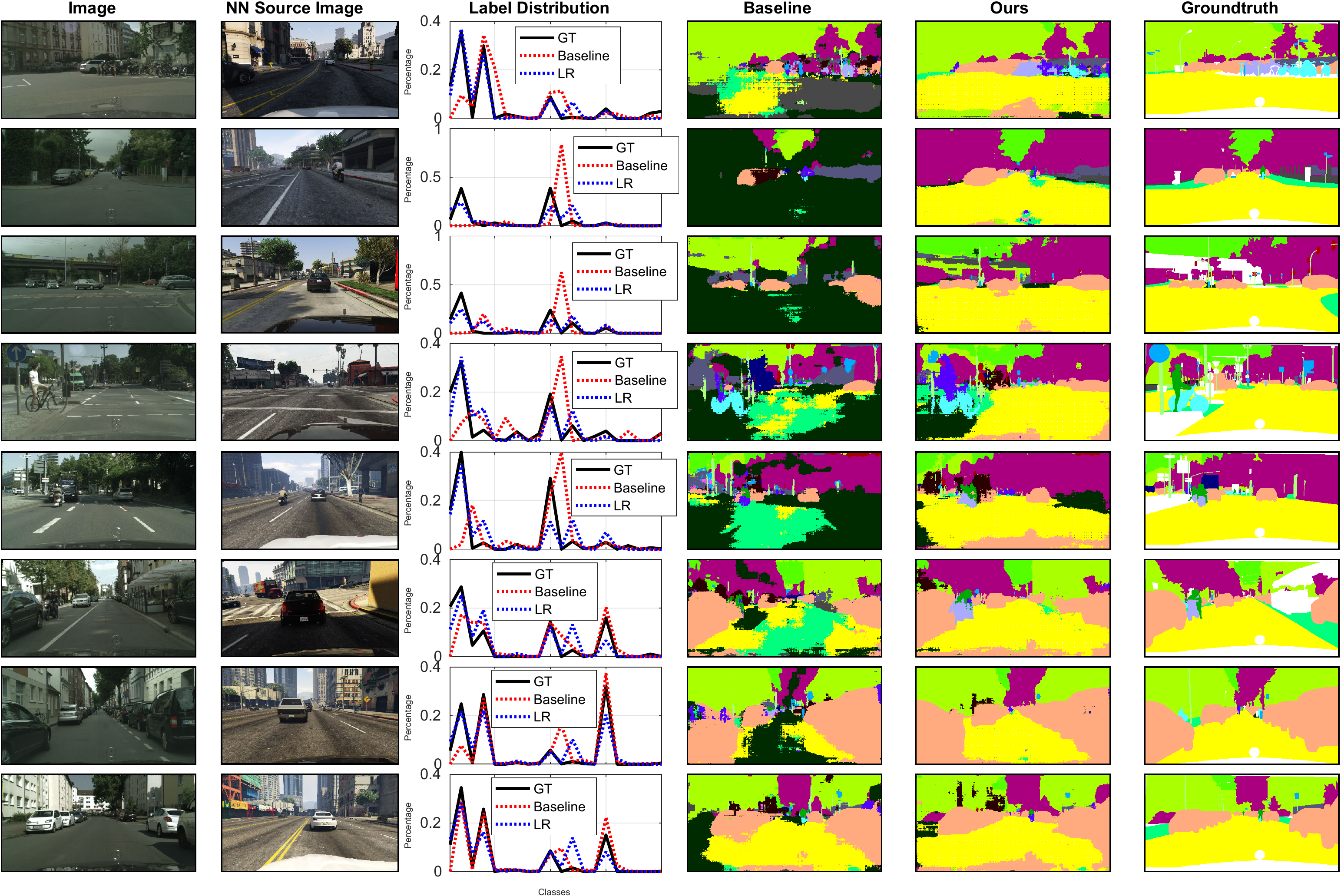}
    \vspace{-20pt}
  \caption{Qualitative semantic segmentation results on the Cityscapes dataset~\cite{ros_synthia_2016} (target domain). For each target image in the first column, we retrieve its nearest neighbor from the GTA~\cite{richter_playing_2016} dataset (source domain). The third column plots the label distributions due to the groundtruth pixel-wise semantic annotation, the predictions by the baseline network with no adaptation, and the inferred distribution by logistic regression. The last three columns are the segmentation results by the baseline network, our domain adaptation approach, and human annotators, respectively.} \label{fGTAQualitative}
  \vspace{-15pt}
\end{sidewaysfigure*}

\section{Experiments} \label{sExperiments}

In this section, we run extensive experiments to verify the effectiveness of our approach and study several variations of it to gain a thorough understanding about how different components contribute to the overall results. We investigate the global image-wise  label distribution and the local landmark superpixels by separate experiments. We also empirically examine the effects of distinct granularities of the superpixels as well as different feature representations of the superpixels.  Moreover, we compare our approach with several competing baselines for the adaptation of various deep neural networks from two synthetic datasets for the  urban scene segmentation task.  We use confusion matrices to show how different classes of the urban scene images intervene with each other. Finally, we run few-shot adaptation experiments by gradually revealing some labels of the images of the target domain. The results show that, even when more than 1,000 target images are labeled (50\% of the Cityscapes training set~\cite{cordts2016cityscapes}), the adaptation from synthetic images is still able to boost the results by a relatively large margin. 

\subsection{Segmentation network and optimization}

In most of our experiments, we use FCN-8s~\cite{long_fully_2015} as our semantic segmentation network. We initialize its convolutional layers with VGG-19~\cite{simonyan_very_2014} and then train it using the AdaDelta optimizer~\cite{zeiler_adadelta:_2012} with default parameters. Each mini-batch contains five source images and five randomly chosen target images. When we train the baseline network with no adaptation, however, we try to use the largest possible mini-batch which includes 15 source images. The network is implemented in Keras~\cite{chollet_keras_2015} and Theano~\cite{al-rfou_theano:_2016}. We train different versions of the network on a single Tesla K40 GPU.

Additionally, we run an ablation experiment using the state-of-the-art segmentation neural network ADEMXAPP~\cite{wu_wider_2016}. The training details and network architecture are described in Section~\ref{sADEMXAPP}.

We note that our curriculum domain adaptation can be readily applied to other segmentation networks (e.g.,~\cite{noh_learning_2015,chen_semantic_2014}). Once we infer the label distributions of the unlabeled target images and some of their landmark superpixels, we can use them to train different segmentation networks by eq.~(\ref{eLoss}) without changing them.


\subsection{Datasets and evaluation}
We use the publicly available \textbf{Cityscapes}~\cite{cordts2016cityscapes} as our target domain in the experiments. For the source domains of synthesized images, we test both \textbf{GTA}~\cite{richter_playing_2016} and \textbf{SYNTHIA}~\cite{ros_synthia_2016}. 

{Cityscapes} is a real-world vehicle-egocentric image dataset collected from 50 cities in Germany and the countries around. It provides four disjoint subsets: 2,993 training images, 503 validation image, 1,531 test images, and 20,021 auxiliary images. All the training, validation, and test images are accurately annotated with per-pixel category labels, and the auxiliary set is coarsely labeled.  There are 34 distinct categories in the dataset. Among them, 19 categories are officially recommended for training and evaluation.

{SYNTHIA}~\cite{ros_synthia_2016} is a large dataset of synthetic images and provides a particular subset, called SYNTHIA-RAND-CITYSCAPES, to pair with {Cityscapes}. This subset contains 9,400  images that are automatically annotated with 12 object categories,  one void class, and some unnamed classes. Note that the virtual city used to generate the synthetic images does not correspond to any of the real cities covered by {Cityscapes}.

{GTA}~\cite{richter_playing_2016} is a synthetic vehicle-egocentric image dataset collected from the open world in a realistically rendered computer game, Grand Theft Auto V (GTA). It contains 24,996 images. Unlike the SYNTHIA dataset, its semantic segmentation annotation is fully compatible with the Cityscapes dataset. Hence, we will use all the 19 official training classes in our experiment.

\subsubsection{Domain idiosyncrasies}
Although all datasets depict urban scenes and both {SYNTHIA} and {GTA} are created to be as photo-realistic as possible, they are mismatched domains in several ways. The most noticeable difference is probably the coarse-grained textures in SYNTHIA; very similar texture patterns repeat in a regular manner across different images. The textures in GTA  are better but still visibly artificial. In contrast, the Cityscapes images are captured by high-quality dash-cameras. Another major distinction is the variability in view angles. Since Cityscapes images are recorded by the dash cameras mounted on a moving car, they are viewed from almost a constant angle that is about parallel to the ground.  More diverse view angles are employed by SYNTHIA --- it seems like some cameras are placed on the buildings that are significantly higher than a bus. Most GTA images are dashcam images, but some of them are captured from the view points of the pedestrians. In addition, some of the SYNTHIA images are severely shadowed by extreme lighting conditions, while we find no such conditions in the Cityscapes images. Finally, there is a subtle difference in  color between the synthetic images and the real ones due to the graphics rendering engines' systematic performance. For instance, the GTA images are overly saturated (cf.\ Figure~\ref{fGTAQualitative}) and SYNTHIA images are overly bright in general. These combined factors, among others, make the domain adaptation from SYNTHIA and GTA to Cityscapes a very challenging problem. 

Figure~\ref{fQualitative} and Figure~\ref{fGTAQualitative} show some example images from the three datasets. We pair each Cityscapes image with its nearest neighbor in SYNTHIA/GTA, retrieved by the Inception-Resnet-v2 \cite{szegedy_inception-v4_2016} features. However, many of the cross-dataset nearest neighbors are visually very different from the query images, verifying the dramatic disparity  between the two domains.



\subsubsection{Evaluation} We use the evaluation code released along with the Cityscapes dataset to evaluate our results. It calculates the PASCAL VOC intersection-over-union, i.e., 
$\text{IoU}=\frac{\text{TP}}{\text{TP}+\text{FP}+\text{FN}}$~\cite{everingham_pascal_2015}, where  TP, FP, and FN are the numbers of true positive, false positive, and false
negative pixels, respectively, determined over the whole test set. Since we have to resize the images before feeding them to the segmentation network, we resize the output segmentation mask back to the original image size before running the evaluation against the groundtruth annotations.


\begin{table}
\centering
\caption{ $\chi^2$ distances between the groundtruth label distributions and those predicted by different methods for the adaptation from SYNTHIA to Cityscapes.}
\label{tLayoutmismatch}
\scalebox{1}{
\begin{tabular}{l|ccccc}
\hline
Method   & Uniform & NoAdapt & Src mean  & NN & \textbf{LR}\\
\hline
$\chi^2$ Distance   & 1.13 & 0.65 & 0.44  & 0.33 & \textbf{0.27}\\
\hline
\end{tabular}
}
\end{table}

\subsection{Results of inferring global label distribution}
Before presenting the final semantic segmentation results, we first compare  different approaches to inferring the global label distributions over the target images (cf.\ Section~\ref{subsLD}). We use SYNTHIA and Cityscapes' held-out validation images as the source domain and the target domain, respectively, in this experiment. 

In Table~\ref{tLayoutmismatch}, we compare the estimated label distributions with the groundtruth ones using the $\chi^2$ distance, the smaller the better. We see that the baseline network (NoAdapt), which is directly learned from the source domain without any adaptation methods, outperforms the dumb uniform distribution (Uniform) and yet no other methods. This confirms that the baseline network gives rise to severely disproportional predictions on the target domain. 

Another dumb prediction (Src mean), i.e., using the mean of  all label distributions over the source domain as the prediction for any target image, however, performs reasonably well mainly because the protocol layouts of the urban scene images. This result implies that the images from the simulation environments share at least similar layouts as the real images,
indicating the potential value of the simulated source domains for the semantic segmentation task of urban scenes. 

Finally, the nearest neighbor (NN) based method and the multinomial logistic regression (LR) (cf.\ Section~\ref{subsLD}) perform the best. We use the output of LR on the target domain in our remaining experiments.

\subsection{Domain adaptation experiments} \label{sDAExp}
In this section, we present our main results of this paper, i.e., comparison results for the domain adaptation from simulation to real images for the semantic segmentation task. Here we focus on the base segmentation neural network FCN-8s~\cite{long_fully_2015}. Another network, ADEMXAPP~\cite{wu_wider_2016}, is studied in Section~\ref{sADEMXAPP}.

Since our ultimate goal is to solve the semantic segmentation problem for the real images of urban scenes, we take Cityscapes as the target domain and SYNTHIA/GTA as the source domain. We split 500 images out of the Cityscapes training set for the validation purpose (e.g., to monitor the convergence of the networks). In training,  we randomly sample mini-matches from both the images and labels of SYNTHIA/GTA and the remaining images of Cityscapes yet with no labels. The original Cityscapes validation set is used as our test set. 

All the 19 classes provided by GTA are used in the experiments. For the adaptation from SYNTHIA to Cityscapes, we manually find 16 common classes between the two datasets: sky, building, road, sidewalk, fence, vegetation, pole, car, traffic sign, person, bicycle, motorcycle, traffic light, bus, wall, and rider. The last four are unnamed and yet labeled in SYNTHIA. 


\subsubsection{Baselines} \label{sBaseline}

We mainly compare our approach to the following competing methods. Section~\ref{SecReview} supplies additional discussions about and comparisons with more related works. 

\begin{description}

\item [No adaptation (NoAdapt).] We directly train the FCN-8s model on the source domain (SYNTHIA or GTA) without applying any domain adaptation methods. This is the most basic baseline in our experiments. 

\item [Superpixel classification (SP).] Recall that we have trained a multi-class SVM using the dominant labels of the superpixels in the source domain. We then use it to classify the target superpixels. 

\item [Landmark superpixels (SP Lndmk).] We keep the top 30\% most confidently classified superpixels  as the landmarks to regularize our segmentation network during training (cf.\ Section~\ref{subsSP}). It is worth examining the classification results of these superpixels. We execute the evaluation after assigning the void class label to the other pixels of the images. In addition to the IoU, we have also evaluated the classification results of the superpixels by accuracy for the domain adaptation experiments from SYNTHIA to Cityscapes. We find that the classification accuracy is 71\% for all the superpixels of the target domain. For the top 30\% landmark superpixels, the classification accuracy is more than 88\%. 


\item [FCNs in the wild (FCN Wld).] Hoffman et al.'s work~{\cite{hoffman_fcns_2016}} was the only existing one addressing the same problem as ours when we published the conference version~\cite{zhang2017curriculum} of this work, to the best of our knowledge. They introduce a pixel-level adversarial loss to the intermediate layers of the network and impose constraints to the network output. Their experimental setup is about identical to ours except that they do not specify which part of Cityscapes is considered as the test set. Nonetheless, we include their results for comparison to put our work in a better perspective. 

\end{description}

\subsubsection{Comparison results}

\begin{table*}
\begin{adjustwidth}{-.5in}{-.5in}  
    \centering
    \small
\caption{Comparison results for adapting the FCN-8s model from SYNTHIA to Cityscapes.}
\label{Tresults_SYNTHIA}
\scalebox{0.92}{
\begin{tabular}{l|c|cccccccccccccccc}

\hline
\multirow{2}{*}[-2em]{ Method~~~~\%} & \multirow{2}{*}[-2em]{ IoU} & \multicolumn{16}{c}{\textbf{SYNTHIA2Cityscapes} Class-wise IoU}\\

 & &  \rot{bike} & \rot{fence} & \rot{wall} & \rot{t-sign} & \rot{pole} & \rot{mbike} & \rot{t-light} & \rot{sky} & \rot{bus} & \rot{rider} & \rot{veg} & \rot{bldg} & \rot{car} & \rot{person} & \rot{sidewalk} & \rot{road}\\
 \hline\hline
 NoAdapt~\cite{hoffman_fcns_2016}  & 17.4 & 0.0 & 0.0 & 1.2 & 7.2 & 15.1 & 0.1 & 0.0 & 66.8 & 3.9 & 1.5 & 30.3 & 29.7 & 47.3 & 51.1 & 17.7 & 6.4\\

FCN Wld~\cite{hoffman_fcns_2016} & 20.2 & 0.6 & 0.0 & \textbf{4.4} & \textbf{11.7} & 20.3 & 0.2 & 0.1 & 68.7 & 3.2 & 3.8 & 42.3 & 30.8 & \textbf{54.0} & \textbf{51.2} & 19.6 & 11.5\\
\hline
NoAdapt & 22.0 & 18.0 & 0.5 & 0.8 & 5.3 & 21.5 & 0.5 & 8.0 & 75.6 & 4.5 & 9.0 & 72.4 & 59.6 & 23.6 & 35.1 & 11.2 & 5.6\\
\textbf{NoAdapt (CC)} & 22.6 & 22.2 & 0.5 & 1.1 & 5.0 & 21.5 & 0.6 & 8.5 & 73.4 & 4.8 & 9.2 & 73.2 & 56.7 & 28.4 & 34.8 & 12.1 & 9.1\\
\hline
\textbf{Ours (I)} & 25.5 & 16.7 & 0.8 & 2.3 & 6.4 & \textbf{21.7} & 1.0 & \textbf{9.9} & 59.6 & 12.1 & 7.9 & 70.2 & 67.5 & 32.0 & 29.3 & 18.1 & 51.9\\
\textbf{Ours (CC+I)} & 27.3 & \textbf{31.2} & \textbf{1.3} & 3.9 & 6.0 & 19.4 & \textbf{2.1} & 9.2 & 61.2 & 11.2 & 7.4 & 68.3 & 65.1 & 41.4 & 29.3 & 18.9 & 60.6\\
\hline
SP Lndmk (CC) & 23.1 & 0.0 & 0.0 & 0.0 & 0.0 & 0.0 & 0.0 & 0.0 & 82.6 & 27.8 & 0.0 & 73.1 & 67.9 & 40.7 & 5.8 & 10.3 & 62.2\\
SP (CC) & 25.6 & 0.0 & 0.0 & 0.0 & 0.0 & 0.0 & 0.0 & 0.0 & 80.1 & 22.7 & 0.0 & 72.2 & 69.7 & 45.6 & 25.0 & 19.4 & 74.8\\

\textbf{Ours (SP)} & 28.1 & 10.2 & 0.4 & 0.1 & 2.7 & 8.1 & 0.8 & 3.7 & 68.7 & 21.4 & 7.9 & 75.5 & 74.6 & 42.9 & 47.3 & 23.9 & 61.8\\
\textbf{Ours (CC+SP)} & 28.9 & 17.7 & 0.5 & 0.5 & 3.4 & 10.9 & 1.8 & 5.4 & 73.4 & 17.6 & 9.9 & 76.8 & 74.5 & 43.7 & 44.4 & 22.4 & 59.6\\
\hline
\textbf{Ours (I+SP)} & 29.0 & 13.1 & 0.5 & 0.1 & 3.0 & 10.7 & 0.7 & 3.7 & 70.6 & 20.7 & 8.2 & 76.1 & \textbf{74.9} & 43.2 & 47.1 & \textbf{26.1} & 65.2\\
\textbf{Ours (CC+I+SP)} & \textbf{\underline{29.7}} & 20.3 & 0.6 & 0.5 & 4.3 & 14.0 & 1.9 & 5.3 & 73.7 & 21.2 & \textbf{11.0} & \textbf{77.8} & 74.7 & 44.8 & 45.0 & 23.1 & 57.4\\
\hline
\end{tabular}
}
\end{adjustwidth}
\end{table*}

\begin{table*}
\begin{adjustwidth}{-.5in}{-.5in}  
    \centering
    \small
\caption{Comparison results for adapting the FCN-8s model from GTA to Cityscapes.}
\label{Tresults_GTA}
\scalebox{0.81}{
\begin{tabular}{l|c|ccccccccccccccccccc}

\hline
\multirow{2}{*}[-2em]{ Method~~~~\%} & \multirow{2}{*}[-2em]{ IoU} & \multicolumn{19}{c}{\textbf{GTA2Cityscapes} Class-wise IoU}\\

 & &  \rot{bike} & \rot{fence} & \rot{wall} & \rot{t-sign} & \rot{pole} & \rot{mbike} & \rot{t-light} & \rot{sky} & \rot{bus} & \rot{rider} & \rot{veg}  & \rot{terrain} & \rot{train} & \rot{bldg} & \rot{car} & \rot{person} & \rot{truck} & \rot{sidewalk} & \rot{road}\\
 \hline\hline
 NoAdapt~\cite{hoffman_fcns_2016}  & 21.1 & 0.0 & 3.1 & 7.4 & 1.0 & 16.0 & 0.0 & 10.4 & 58.9 & 3.7 & 1.0 & 76.5 & 13 & 0.0 & 47.7 & 67.1 & 36 & 9.5 & 18.9 & 31.9\\

FCN Wld~\cite{hoffman_fcns_2016} & 27.1 & 0.0 & 5.4 & \textbf{14.9} & 2.7 & 10.9 & 3.5 & 14.2 & 64.6 & 7.3 & 4.2 & 79.2 & 21.3 & 0.0 & 62.1 & \textbf{70.4} & \textbf{44.1} & 8.0 & \textbf{32.4} & 70.4\\

\hline
NoAdapt & 22.3 & 13.8 & 8.7 & 7.3 & 16.8 & 21.0 & 4.3 & 14.9 & 64.4 & 5.0 & 17.5 & 45.9 & 2.4 & 6.9 & 64.1 & 55.3 & 41.6 & 8.4 & 6.8 & 18.1\\

\textbf{NoAdapt (CC)} & 26.2 & \textbf{16.2} & 10.9 & 8.8 & \textbf{18.5} & \textbf{23.3} & 7.0 & 13.2 & 62.7 & 5.4 & \textbf{19.0} & 65.1 & 5.8 & 2.3 & 64.8 & 63.9 & 42.2 & 9.2 & 13.8 & 45.0\\
\hline
\textbf{Ours (I)} & 23.1 & 9.5 & 9.4 & 10.2 & 14.0 & 20.2 & 3.8 & 13.6 & 63.8 & 3.4 & 10.6 & 56.9 & 2.8 & \textbf{10.9} & 69.7 & 60.5 & 31.8 & 10.9 & 10.8 & 26.4\\
\textbf{Ours (CC+I)} & 28.5 & 7.2 & 9.4 & 11.1 & 13.4 & 23.1 & 9.6 & 15.1 & 64.6 & 5.9 & 15.5 & 71.1 & 10.3 & 3.9 & 67.7 & 62.3 & 43.0 & 14.0 & 23.0 & 71.6\\
\hline
SP Lndmk (CC) & 21.6 & 0.0 & 0.0 & 0.0 & 0.0 & 0.0 & 0.0 & 0.0 & 82.4 & 9.1 & 0.0 & 74.4 & 22.2 & 0.0 & 70.3 & 53.1 & 15.3 & 11.2 & 6.9 & 65.8\\

SP (CC) & 26.8 & 0.3 & 2.3 & 6.8 & 0.0 & 0.2 & 3.4 & 0.0 & 80.5 & 25.5 & 4.1 & 73.5 & 31.4 & 0.0 & 71.0 & 61.6 & 28.2 & 30.4 & 17.3 & 73.3\\

\textbf{Ours (SP)} & 27.8 & 15.6 & 11.7 & 5.7 & 12.0 & 9.2 & 12.9 & 15.5 & 64.9 & 15.5 & 9.1 & 74.6 & 11.1 & 0.0 & 70.5 & 56.1 & 34.8 & 15.9 & 21.8 & 72.1\\

\textbf{Ours (CC+SP)} & 30.2 & 10.4 & \textbf{13.6} & 10.3 & 14.0 & 13.9 & 18.8 & 16.5 & 73.6 & 14.1 & 9.5 & 79.2 & 12.9 & 0.0 & 74.3 & 63.5 & 33.1 & 18.9 & 27.5 & 70.5\\
\hline
\textbf{Ours (I+SP)} & 28.9 & 14.6 & 11.9 & 6.0 & 11.1 & 8.4 & 16.8 & 16.3 & 66.5 & 18.9 & 9.3 & 75.7 & 13.3 & 0.0 & 71.7 & 55.2 & 38.0 & 18.8 & 22.0 & \textbf{74.9}\\
\textbf{Ours (CC+I+SP)} & \textbf{\underline{31.4}} & 12.0 & 13.2 & 12.1 & 14.1 & 15.3 & \textbf{19.3} & \textbf{16.8} & 75.5 & 19.0 & 10.0 & \textbf{79.3} & 14.5 & 0.0 & \textbf{74.9} & 62.1 & 35.7 & 20.6 & 30.0 & 72.9\\

\hline
\end{tabular}
}
\end{adjustwidth}
\end{table*}

The overall comparison results of adaptating from SYNTHIA to Cityscapes (SYNTHIA2Cityscapes) are shown in Table~\ref{Tresults_SYNTHIA}. We also present the results of adapting from  GTA to Cityscapes (GTA2Cityscapes) in Table~\ref{Tresults_GTA}. Immediately, we note that all our domain adaptation results are significantly better than those without adaptation (NoAdapt) in both tables. 

We denote by (\textbf{Ours (I)}) the network regularized by the global label distributions over the target images. Although one may wonder that the image-wise label distributions are too abstract to supervise the pixel-wise discriminative network, the gain is actually significant. They are able to correct some obvious errors of the baseline network, such as the disproportional predictions about road and sidewalk (cf.\ the results of \textbf{Ours (I)} vs.\ NoAdapt in the last two columns of Tables~\ref{Tresults_SYNTHIA} and \ref{Tresults_GTA}).

It is interesting to see that both superpixel classification-based segmentation results (SP and SP Lndmk) are also better than the baseline network (NoAdapt). The label distributions obtained over the landmark superpixels boost the segmentation network (\textbf{Ours (SP)}) to the mean IoU of 28.1\% and 27.8\% respectively when adapting from SYNTHIA and GTA, which are better than those by either superpixel classification or the baseline network individually. We have also tried to use the label distributions over all the superpixels to train the network, and observe little improvement. This is probably because it is too forceful to regularize the network output at every single superpixel especially when the estimated label distributions are not accurate enough.

The superpixel-based methods, including \textbf{Ours (SP)}, miss small objects, such as pole and traffic signs (t-sign), and instead are very accurate for categories like the sky, road, and building, which typically occupy larger image regions. On the contrary, the label distributions on the images give rise to a network (\textbf{Ours (I)}) that performs better on the small objects than \textbf{Ours (SP)}. In other words, they mutually complement to some extent. Re-training the network by using the label distributions over both global images and local landmark superpixels (\textbf{Ours (I+SP)}), we achieve  semantic segmentation results on the target domain that are superior over using either to regularize the network.

Finally, we report the results of our method and its ablated versions (i.e., \textbf{Ours (I+SP)}, \textbf{Ours (I)}, and \textbf{Ours (SP)}) after we apply color constancy to the images (accordingly, the methods are denoted by \textbf{Ours (CC+I+SP)}, \textbf{Ours (CC+I)}, and \textbf{Ours (CC+SP)}). We observe improvements of various degrees over those before the color constancy. Especially, the best results are obtained after we apply the color constancy for adapting from both SYNTHIA and GTA. 


\paragraph{Comparison with FCNs in the wild~\cite{hoffman_fcns_2016}} 
Although we use the same segmentation network (FCN-8s) as~\cite{hoffman_fcns_2016},  our baseline results (NoAdapt) are better than those reported in~\cite{hoffman_fcns_2016}. This may be due to subtle differences in terms of implementation or experimental setup. For both SYNTHIA2Cityscapes and GTA2Cityscapes, we gain larger improvements (7.7\% and 9\%) over the baseline~\cite{hoffman_fcns_2016}. 

\paragraph{Comparison with learning domain-invariant features} At our first attempt to solve the domain adaptation problem for the semantic segmentation of urban scenes, we tried to learn domain invariant features following the deep domain adaptation method~\cite{LongICML15Learning} for classification. In particular, we impose the maximum mean discrepancy~\cite{gretton2012kernel}  over the layer before the output. We name such network layer the feature layer. Since there are virtually three output layers in FCN-8s, we experiment with all the three feature layers correspondingly. We have also tested the domain adaptation by reversing the gradients of a domain classifier~\cite{GaninICML15Unsupervised}. However, none of these efforts lead to any noticeable gain over the baseline network so the results are omitted. 

\subsubsection{ Confusion between classes}

\begin{figure*}
\centering
\begin{tabular}{@{}c@{}c@{}}
  \includegraphics[scale=0.83]{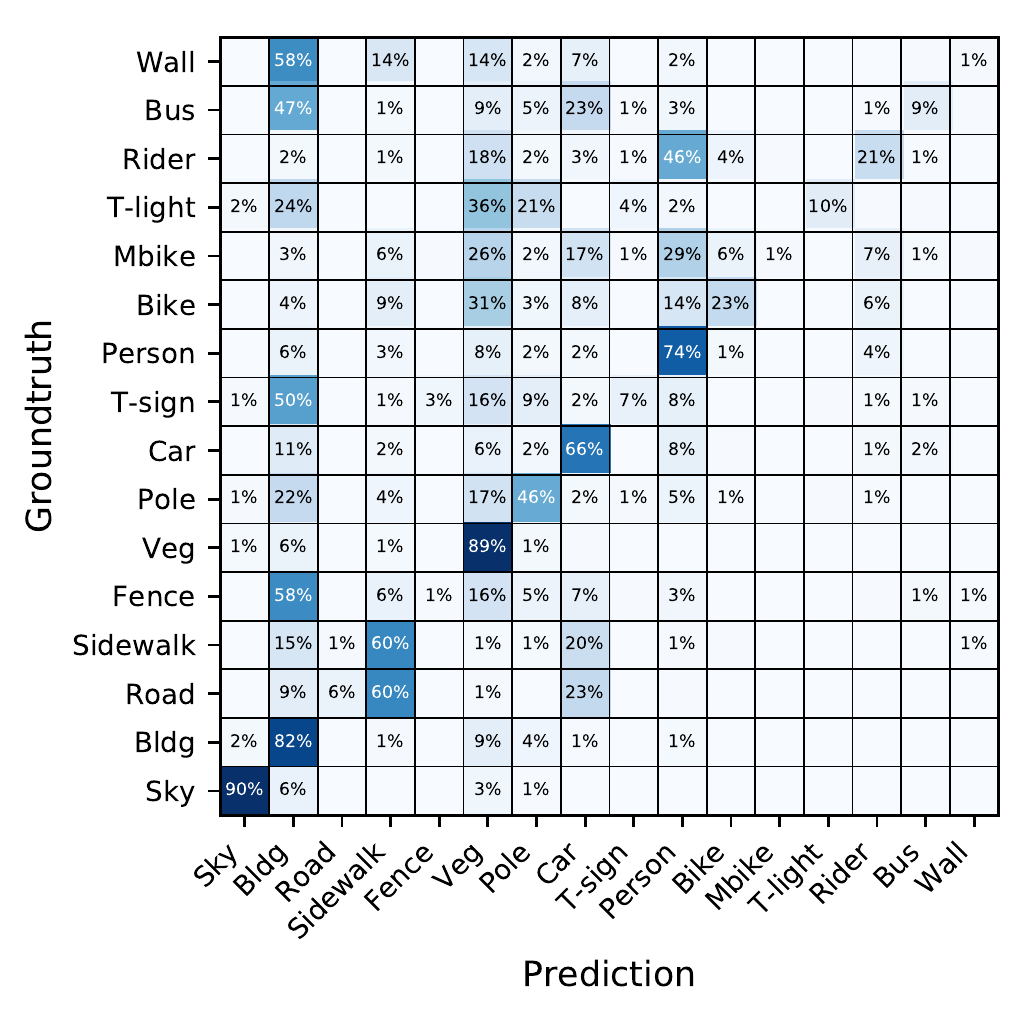} &   \includegraphics[scale=0.83]{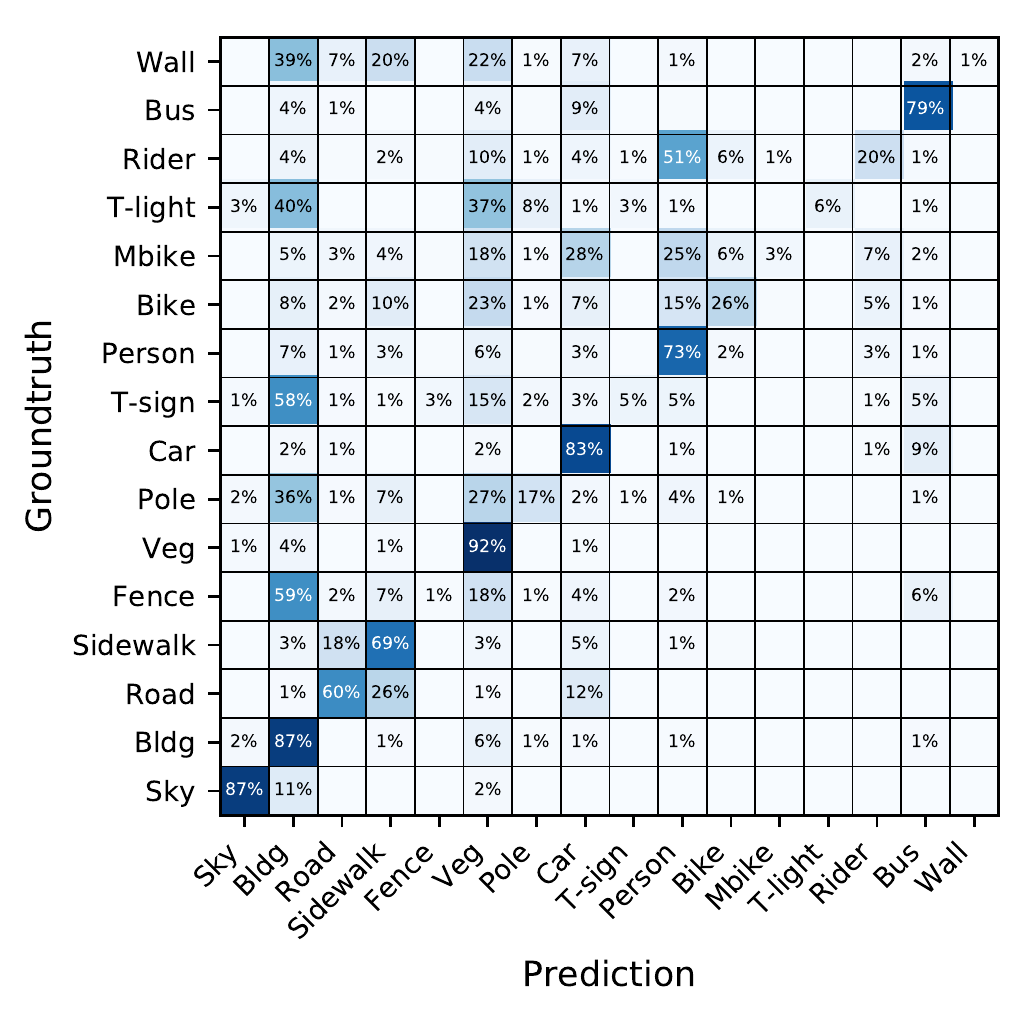} \\
SYNTHIA2Cityscapes baseline & SYNTHIA2Cityscapes \textbf{Ours} \\[8pt]
 \includegraphics[scale=0.83]{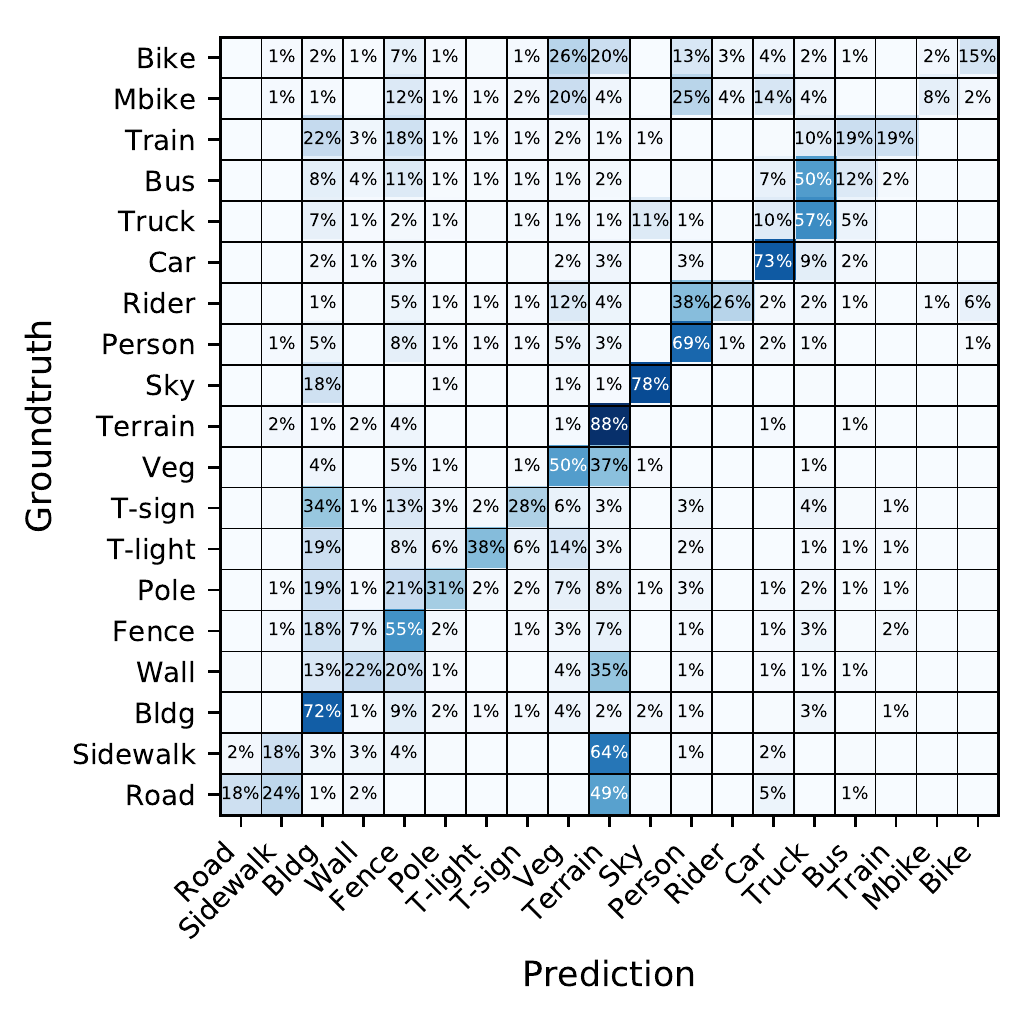} &   \includegraphics[scale=0.83]{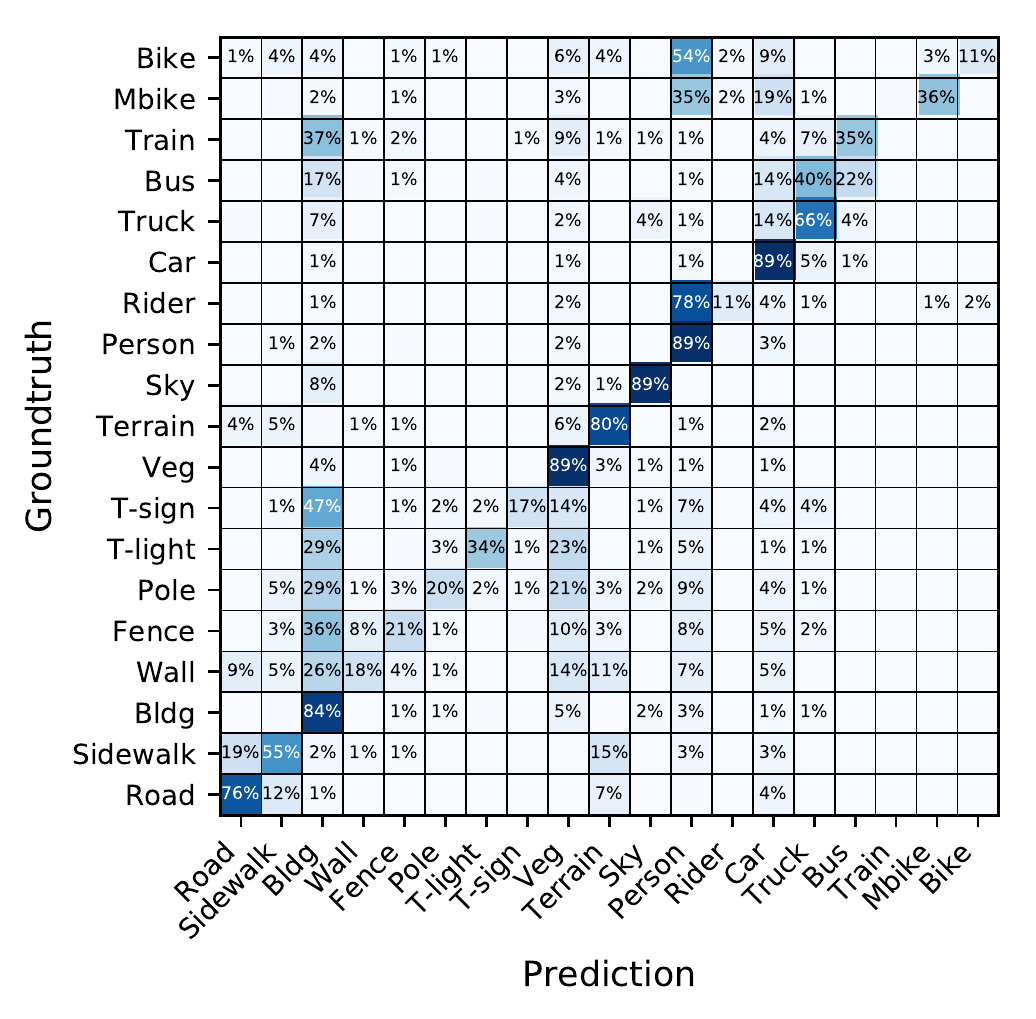} \\
GTA2Cityscapes baseline & GTA2Cityscapes \textbf{Ours} \\
\end{tabular}
\caption{Confusion matrices for the baseline of no adaptation (left) and \textbf{Ours (CC+I+SP)} (right) for the experiments of SYNTHIA-to-Cityscapes (SYNTHIA2Cityscapes, top) and GTA-to-Cityscapes (GTA2Cityscapes, bottom).}
\label{conf_mat_visual}
\end{figure*}

While Tables~\ref{Tresults_SYNTHIA} and \ref{Tresults_GTA} show the overall and per-class results, they do not tell the confusion between different classes. In this section, we provide the confusion matrices of some methods in order to provide more informative analyses about the results. Considering the page limit, we  present in Figure~\ref{conf_mat_visual} the confusion matrices  of NoAdapt and \textbf{Ours (CC+I+SP)} for SYNTHIA2Cityscapes and GTA2Cityscapes. 


We find that a lot of objects are misclassified to the ``building'' category, especially the classes ``pole'', ``traffic sign'', ``traffic light'', ``fence'' and ``wall''. It is probably because those classes often show up beside buildings and they all have huge intra-class variability. Moreover, the ``pole'', ``traffic sign'', and others are very small objects comparing to the ``building''. Some special care is required to disentangle these classes from the ``building'' in the future work. 

Another noticeable confusion is between the ``train'' and the ``bus''. After analyzing the data, we find that this is likely due to the lack of discrimination between the two classes by the datasets themselves, rather than the algorithms. In Figure~\ref{GTA_CS_class_im}, we visualize some trains and buses in the GTA dataset and some trains in the Cityscapes dataset. The difference between the trains and the buses turns out very subtle. We humans could make mistakes too if we do not pay attention to the rails (e.g., the train on the bottom right could be easily misclassified as a bus). Probably this confusion between trains and buses could be alleviated if more training examples can be supplied.



\begin{table*}
\centering
\begin{tabular}{c c c c}
  \vbox to 50pt{\hbox{
  \shortstack{``Train'' \\ GTA}
  }\vss}
  &
  \frame{\includegraphics[width=5cm]{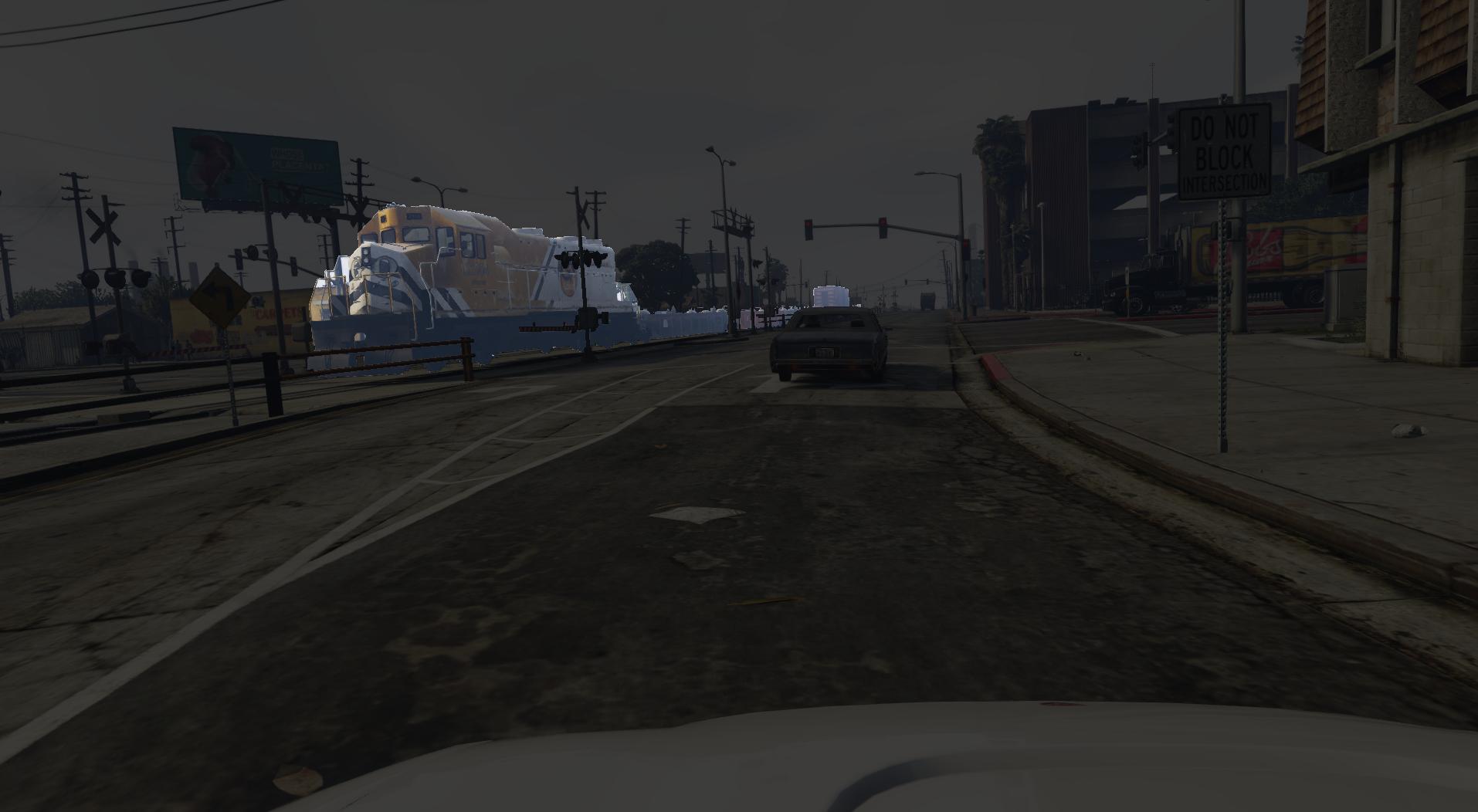}} &  
  \frame{\includegraphics[width=5cm]{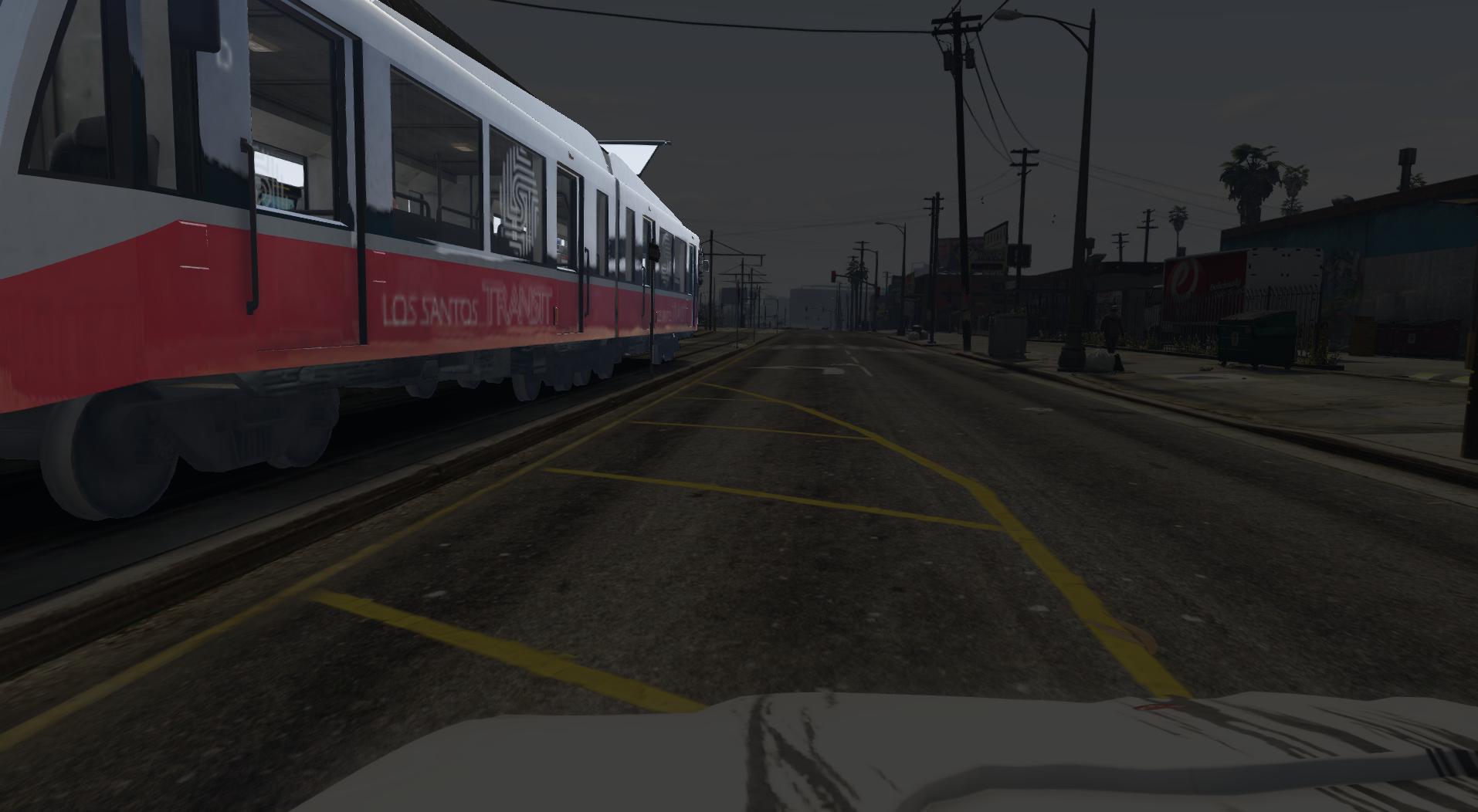}} &  
  \frame{\includegraphics[width=5cm]{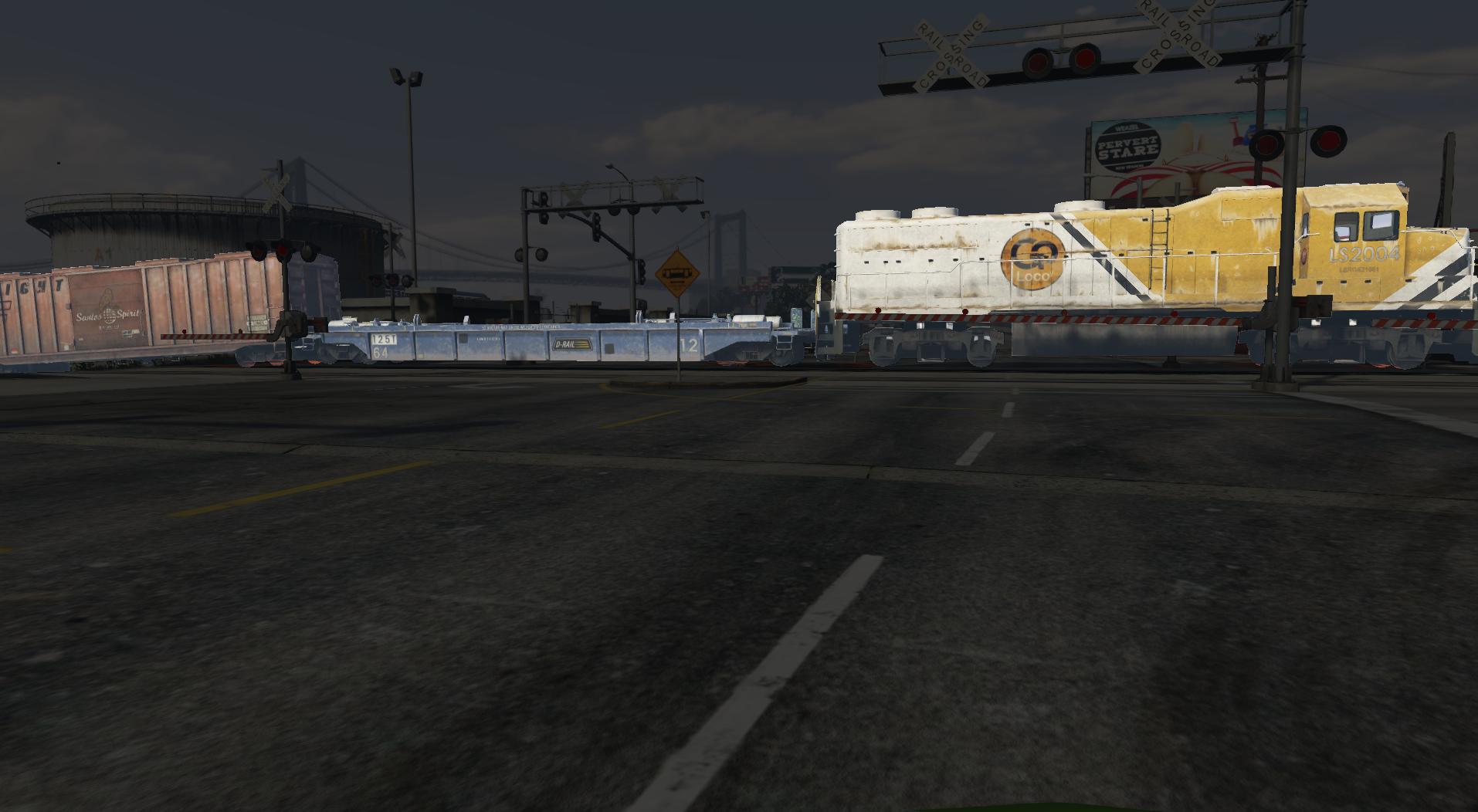}} \\
  \vbox to 50pt{\hbox{
  \shortstack{``Bus'' \\ GTA}
  }\vss}
  &
  \frame{\includegraphics[width=5cm]{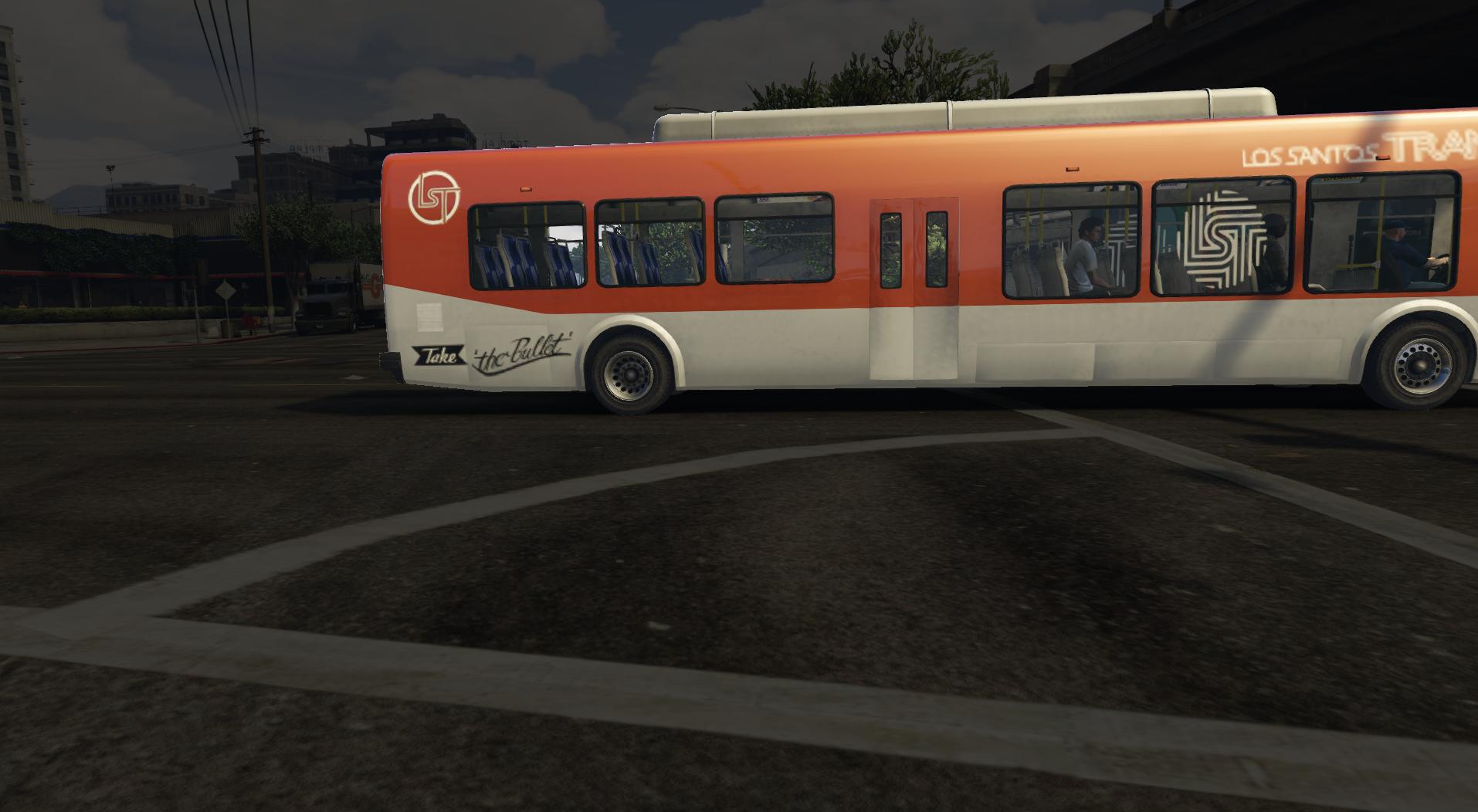}} &  
  \frame{\includegraphics[width=5cm]{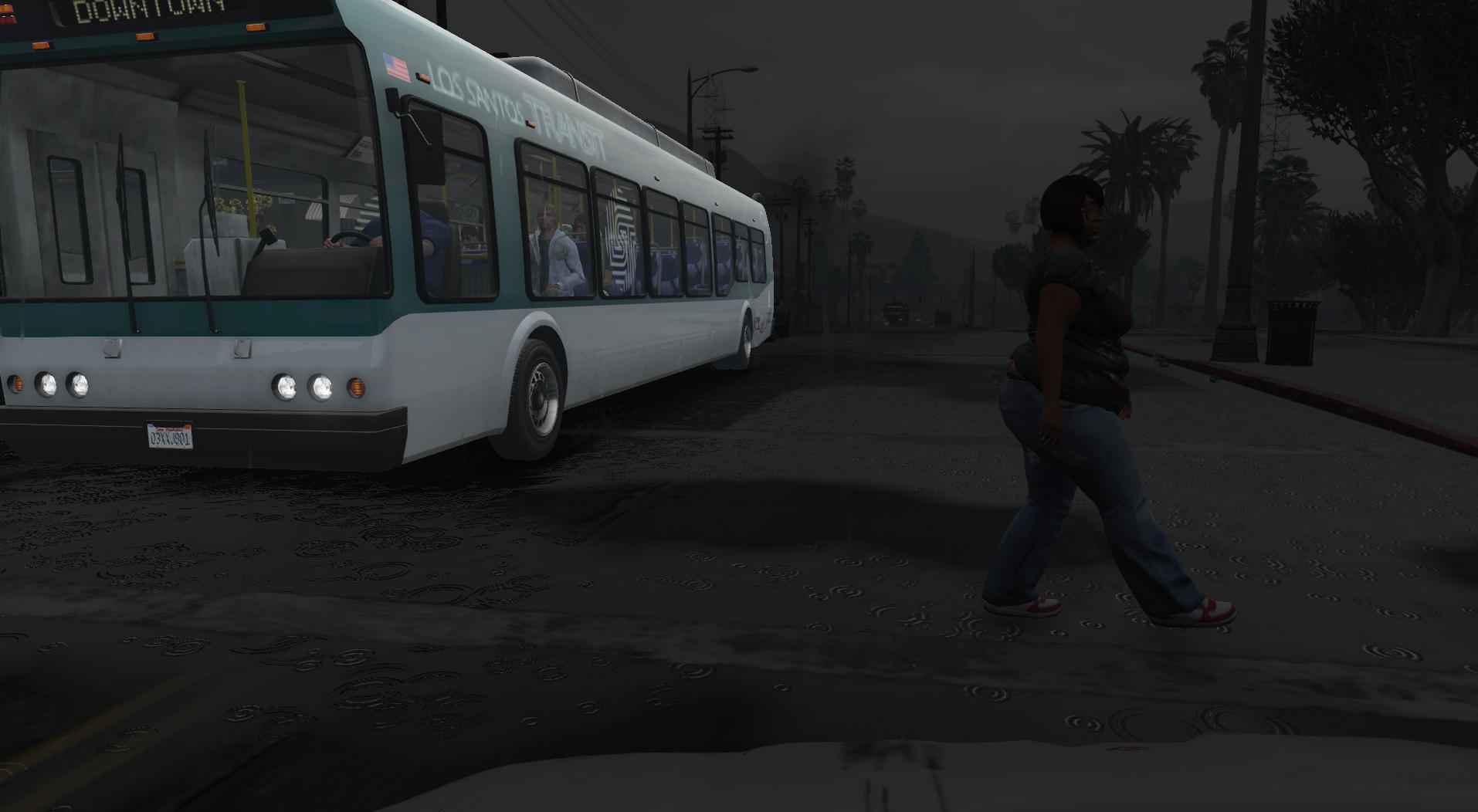}} &  
  \frame{\includegraphics[width=5cm]{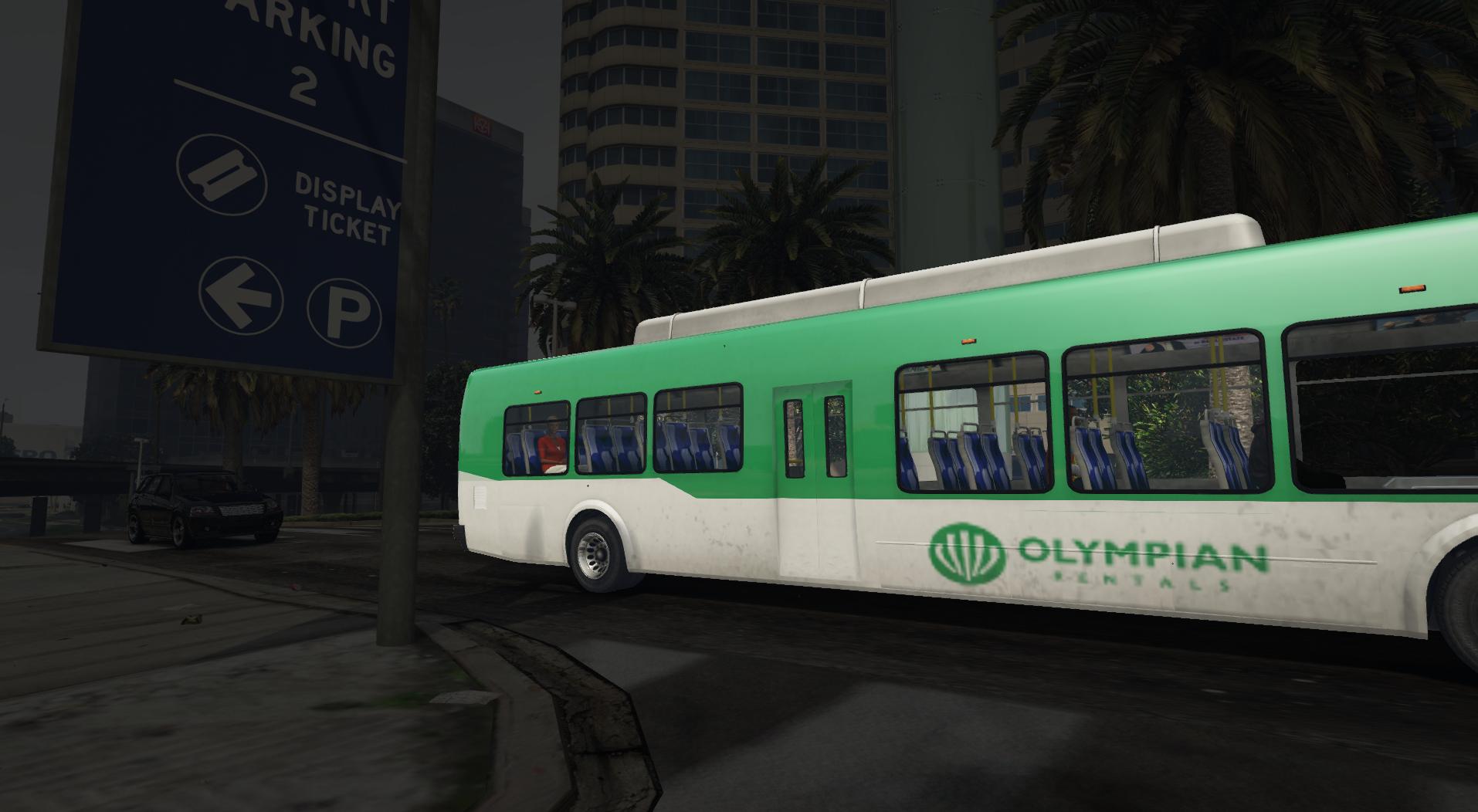}} \\
  \vbox to 50pt{\hbox{
  \shortstack{``Train'' \\ Cityscapes}
  }\vss}
  &
  \frame{\includegraphics[width=5cm]{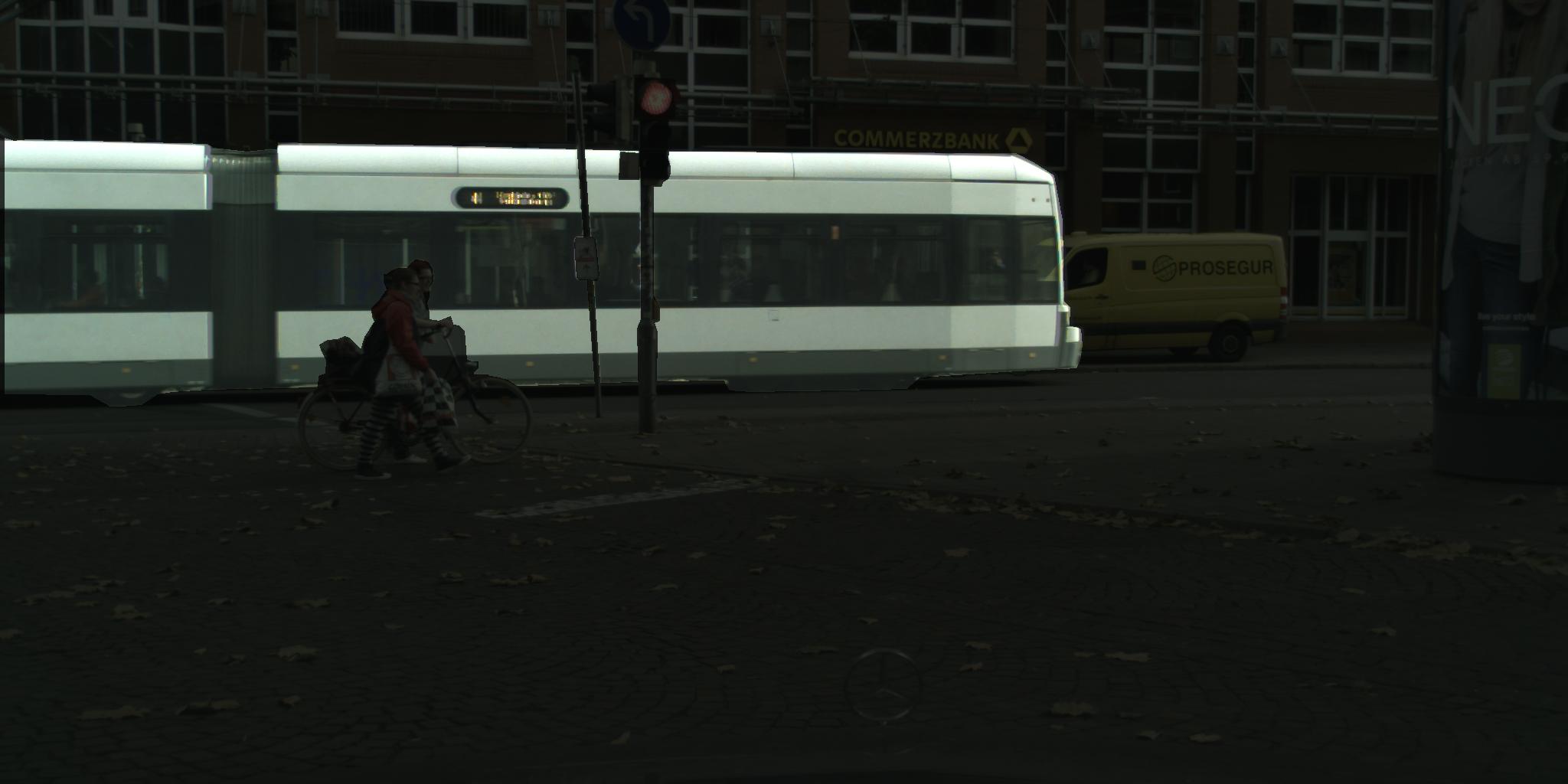}} &  
  \frame{\includegraphics[width=5cm]{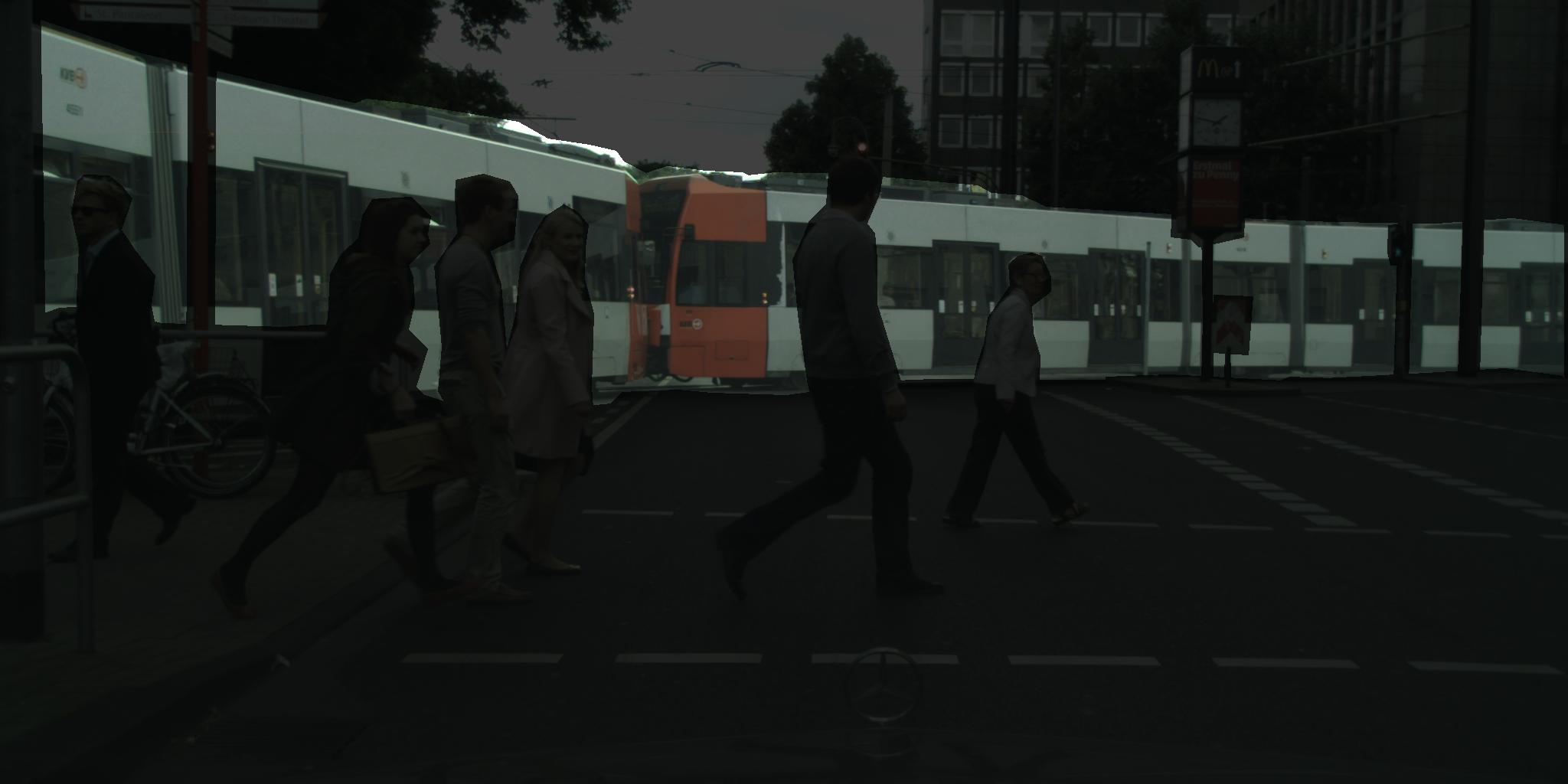}} &  
  \frame{\includegraphics[width=5cm]{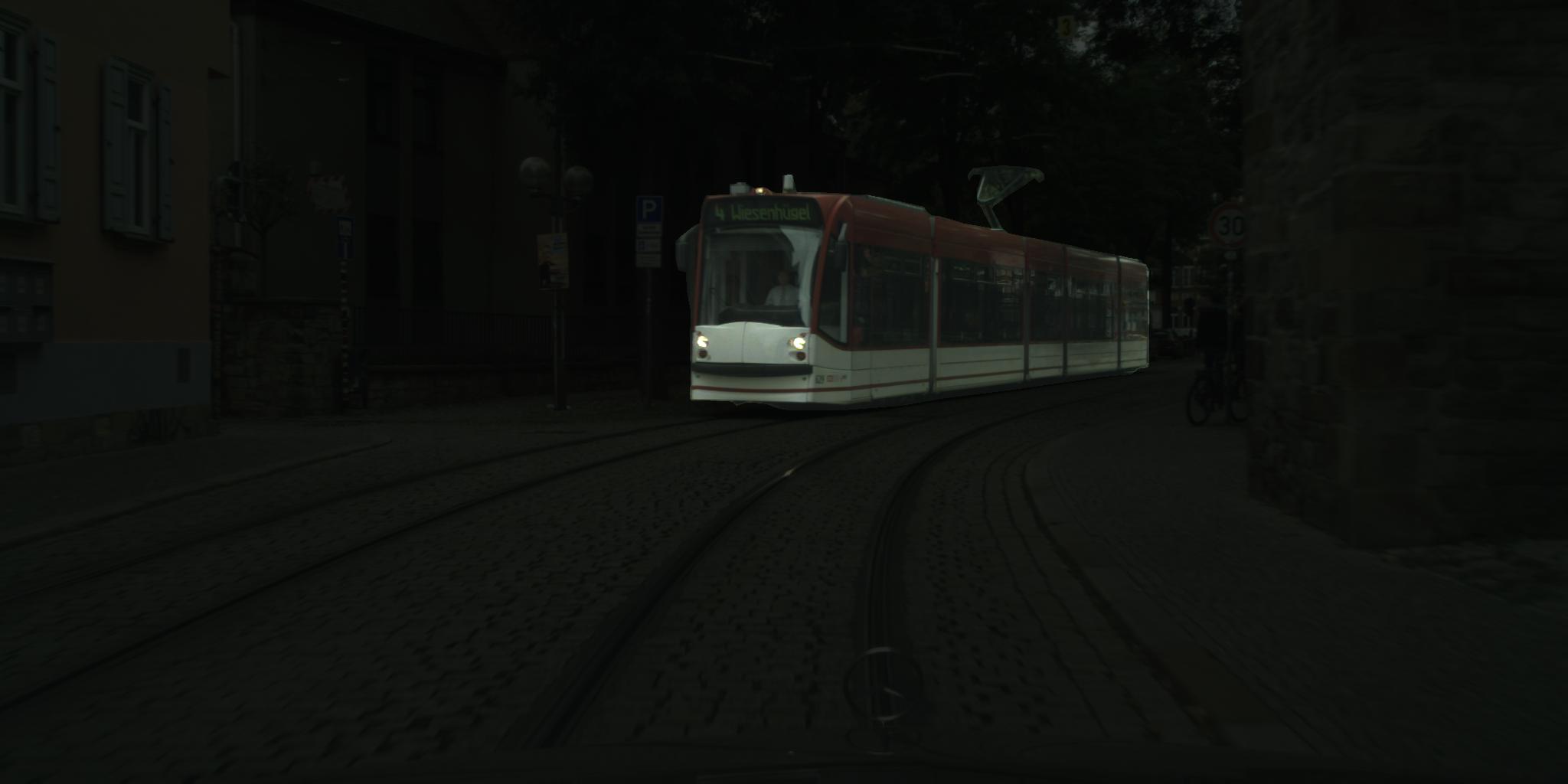}} \\
\end{tabular}
\caption{Some ``train'' and ``bus'' images from the Cityscapes and GTA datasets. We can see that the Cityscapes ``trains'' are more visually similar to the GTA ``buses'' instead of the GTA ``trains''.}
\label{GTA_CS_class_im}
\end{table*}


\begin{table*}
\begin{adjustwidth}{-.5in}{-.5in}  
    \centering
    \small
\caption{Results for the adaptation of FCN-8s from GTA to Cityscapes when we use handcrafted features instead of the CNN features.}
\label{HCresults_GTA}
\scalebox{0.78}{
\begin{tabular}{l|c|ccccccccccccccccccc}

\hline
\multirow{2}{*}[-2em]{ Method~~~~\%} & \multirow{2}{*}[-2em]{ IoU} & \multicolumn{19}{c}{\textbf{GTA2Cityscapes} Class-wise IoU}\\

 & &  \rot{bike} & \rot{fence} & \rot{wall} & \rot{t-sign} & \rot{pole} & \rot{mbike} & \rot{t-light} & \rot{sky} & \rot{bus} & \rot{rider} & \rot{veg}  & \rot{terrain} & \rot{train} & \rot{bldg} & \rot{car} & \rot{person} & \rot{truck} & \rot{sidewalk} & \rot{road}\\
 \hline\hline
 
\textbf{NoAdapt (CC)} & 26.2 &\textbf{ 16.2 }& 10.9 & 8.8 & 18.5 & 23.3 &\textbf{ 7.0 }& 13.2 & 62.7 &\textbf{ 5.4 }&\textbf{ 19.0 }& 65.1 & 5.8 & 2.3 & 64.8 & 63.9 & 42.2 & 9.2 & 13.8 & 45.0\\ 
\hline
\textbf{Ours (CC+BOW)} &  27.9 & 13.8 &\textbf{ 14.0 }& 9.6 & 17.9 & 23.9 & 6.4 & 16.7 & 64.6 & 3.0 & 18.0 & 69.1 & 7.0 & 2.4 & 69.2 & 60.1 &\textbf{ 44.0 }& 10.7 &\textbf{ 19.1 }& 60.8\\ 
\textbf{Ours (CC+FV)} &  28.1 & 13.3 & 10.5 &\textbf{ 12.8 }&\textbf{ 18.6 }& 24.4 & 5.1 & 10.8 & 63.5 & 1.7 & 14.6 &\textbf{ 73.5 }&\textbf{ 10.0 }& 0.3 &\textbf{ 71.6 }&\textbf{ 66.6 }& 40.8 & 6.1 & 11.5 &\textbf{ 79.0}\\ 
\hline
\textbf{Ours (CC+BOW+FV)} &\textbf{\underline{  28.3 }}& 15.3 & 13.3 & 11.6 & 18.5 &\textbf{ 25.1 }& 6.7 &\textbf{ 16.8 }&\textbf{ 66.5 }& 2.9 & 18.4 & 72.2 & 8.7 &\textbf{ 2.6 }& 70.0 & 59.4 & 43.9 &\textbf{ 10.8 }& 19.1 & 56.8\\ 

\hline
\end{tabular}
}
\end{adjustwidth}
\end{table*}


\subsection{Representations of the superpixels}
One may wonder how the representations of the superpixels could change the overall domain adaptation results. We conduct detailed  analyses in this section to reveal some insights about this question. Our main results (Tables~\ref{Tresults_SYNTHIA} and \ref{Tresults_GTA}) are obtained by representing the landmark superpixels with the networks pre-trained on PASCAL CONTEXT~\cite{mottaghi_role_2014}. We are interested in examining whether or not such high-level semantic representations of the superpixels are necessary. Hence, we compare the high-level superpixel descriptors with the following low-level features.

\subsubsection{Handcrafted feature}

\begin{description}

\item [BOW.] We  encode the superpixels with the bag-of-words (BOW)-SIFT features. We first extract dense SIFT features~\cite{lowe2004distinctive} from the input image and then encode those of each superpixel into a 100D BOW vector. The dictionary for the encoding is obtained by K-means clustering. 

\item [FV.] We  also replace the image-level CNN features with the Fisher vectors (FV)~\cite{PerronninECCV10Improving,perronnin2007fisher} for estimating the label distribution of an target image. FV encodes the SIFT features per image into a fixed-dimensional descriptor through a Gaussian mixture model, which has 8 components in this work. An image is then represented by a 2048D vector.  We train the dictionary for BOW and the Gaussian mixture model for FV using the SIFT features of the  GTA dataset only. 

\end{description}

Table~\ref{HCresults_GTA} shows the GTA2Cityscapes results using the  handcrafted features. We denote the resulting methods respectively by  \textbf{Ours (CC+BOW)},  \textbf{Ours (CC+FV)}, and \textbf{Ours (CC+BOW+FV)}. It is interesting to see all of them outperform the baseline \textbf{NoAdapt (CC)}, indicating that our curriculum domain adaptation method is able to leverage the handcrafted features as well. 

\colorchunk{\subsubsection{VGG feature}\label{SecExinfor}
We also test the VGG features~\cite{simonyan_very_2014} of the \texttt{block5\_conv4} and \texttt{block4\_conv4} layers. The network is pre-trained on ImageNet and does not bring in any extra knowledge as the segmentation networks are also pre-trained with ImageNet. In order to extract the features for each superpixel, we average-pool the activations per channel within the superpixel. 

With the new VGG features, the superpixel classification accuracy on the Cityscapes validation set is 76\%, a 5\% boost from the accuracy due to the PASCAL-CONTEXT features. Moreover, the top 30\%, which is the landmark superpixels used in our experiments, is labeled with up to 93\% accuracy (vs.\ 88\% with the PASCAL-CONTEXT features). Thanks to the boost in the classification accurcy of the landmark superpixels, we also observe an 0.6\% gain in mIoU (from 29\% to 29.6\%) on the domain adaptation from SYNTHIA to Cityscapes. These results imply that the representations of the superpixels do influence the final results, but the representations via an extra knowledge base are not necessarily advantageous; the final results with the ImageNet-pretrained VGG features are superior over those with the PASCAL-CONTEXT features.}

\subsection{Granularity of the superpixels}
In this section, we study what is a proper granularity of the superpixels. Intuitively, small superpixels are fine-grained and precisely tracks object boundaries. However, they are less discriminative as a result. What is a proper granularity of the superpixels? How sensitive could the results be to the granularity? To quantitatively answer these questions, we vary the number of superpixels per image to examine their effects on the semantic segmentation results.  The adaptation from GTA of the \textbf{SP (CC)} method, described in Section~\ref{sBaseline}, is reported in Table~\ref{SPNresults_GTA} with various numbers of superpixels per image. In general, the performance increases as the number of superpixels grows until it reaches 300 per image. 

\begin{figure}
    \includegraphics[scale=0.6]{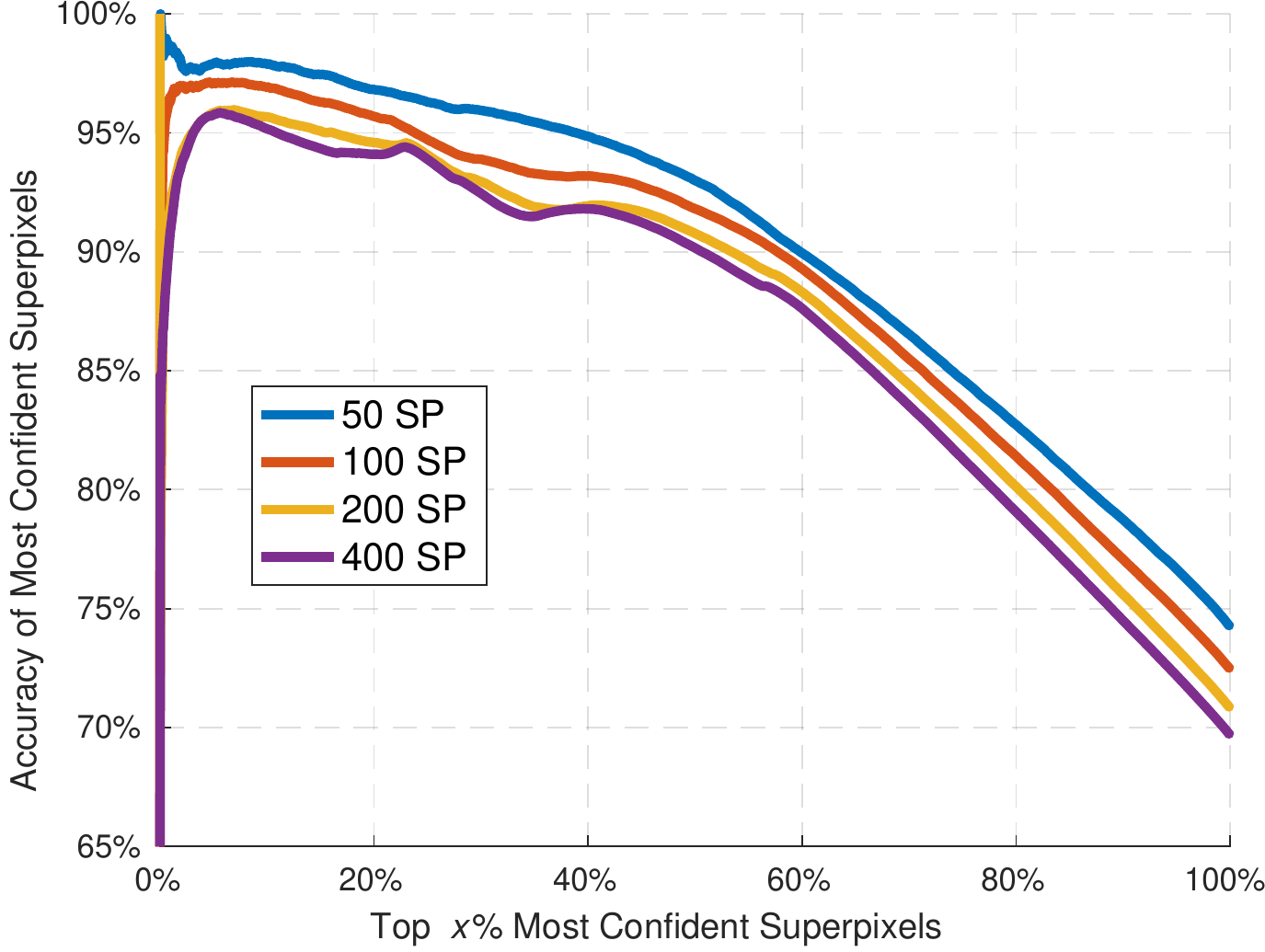}
  \caption{We evaluate how many superpixels are accurate in the top \textit{x}\% confidently predicted superpixels. The experiments are conducted on the validation set of Cityscapes with color constancy.} \label{fSPnumber}
\end{figure}

Besides, we also presented the classification accuracy of the top $x\%$ superpixels in Figure~\ref{fSPnumber}, where $x=0,20,\cdots,100$. We can see that the accuracy is always more than 90\% when we keep the top 20\% or 40\% superpixels per image --- in our experiments, we keep top 30\%. Besides, the accuracy of keep all the superpixels (top 100\%) are not very high, indicating that it is not a good idea to use all superpixels to guide the training of the neural networks.  

\begin{table*}
\begin{adjustwidth}{-.0in}{-.0in}  
    \centering
    \small
\caption{Results for the adaptation of FCN-8s from GTA to Cityscapes when we use different numbers of superpixels per image. Here the images are pre-processed with color constancy.}
\label{SPNresults_GTA}
\scalebox{0.85}{
\begin{tabular}{l|c|ccccccccccccccccccc}

\hline
\multirow{2}{*}[-2em]{ SP \# per image} & \multirow{2}{*}[-2em]{ IoU} & \multicolumn{19}{c}{\textbf{GTA2Cityscapes} Class-wise IoU}\\

 & &  \rot{bike} & \rot{fence} & \rot{wall} & \rot{t-sign} & \rot{pole} & \rot{mbike} & \rot{t-light} & \rot{sky} & \rot{bus} & \rot{rider} & \rot{veg}  & \rot{terrain} & \rot{train} & \rot{bldg} & \rot{car} & \rot{person} & \rot{truck} & \rot{sidewalk} & \rot{road}\\
 \hline\hline
 
50 & 22.9  & 0.0  & 0.0  & 1.5  & 0.0  & 0.0  & 0.0  & 0.0  & 80.1  & 21.9  & 0.0  & 69.3  & 26.2  & 0.0  & 66.3  & 54.9  & 9.9  & 18.3  & 12.6  & \textbf{74.6} \\
\hline
100 &  26.8 & 0.3 & 2.3 & 6.8 & 0.0 & 0.2 & \textbf{3.4} & 0.0 & 80.5 & \textbf{25.5} & 4.1 & 73.5 & 31.4 & 0.0 & 71.0 & 61.6 & 28.2 & \textbf{30.4} & 17.3 & 73.3\\
\hline
200 & 27.3  & 0.2  & 2.2  & 7.2  & 0.0  & 0.8  & 3.0  & 0.0  & 80.5  & 24.4  & \textbf{3.8}  & 75.9  & 32.9  & 0.0  & 72.5  & 63.7  & 31.1  & 27.9  & 18.4  & 74.2 \\
\hline
400 &  \textbf{\underline{27.4}}  & \textbf{0.4}  & \textbf{3.8}  & \textbf{7.2}  & \textbf{0.0}  & \textbf{1.1}  & 1.7  & \textbf{0.0}  & \textbf{80.7}  & 23.2  & 3.7  & \textbf{76.9}  & \textbf{33.3}  & 0.0  & \textbf{72.5}  & \textbf{63.8}  & \textbf{33.1}  & 26.9  & \textbf{18.4}  & 73.3 \\

\hline
\end{tabular}
}
\end{adjustwidth}
\end{table*}

\subsection{Domain adaptation experiments using ADEMXAPP}\label{sADEMXAPP}
Our approach is agnostic to the base semantic segmentation neural networks. In this section, we further investigate a more recent network, ADEMXAPP~\cite{wu_wider_2016}, which is among the few top performing methods on the Cityscapes challenge board. Our experiment setup in this section resembles that of Section~\ref{sDAExp} except that we replace FCN-8s with the ADEMXAPP net. In particular, we reimplement the A1 model of ADEMXAPP 
using the Theano-Keras framework. However, we remove the batch normalization layers in our implementation due to their extensive GPU memory consumption. We follow the authors' suggestions otherwise and initialize the network with the weights pre-trained on Imagenet. We set the  size of the mini-batch to six, three images from the source domain and the other three from the target domain. 

Table~\ref{ADEMXAPPresults_GTA} shows the comparison results for the ADEMXAPP net. We can see it indeed achieves much better results than FCN-8s in general. Nonetheless, the relative trend of our approach against the others remains the same for this ADEMXAPP net. 


\begin{table*}
\begin{adjustwidth}{-.0in}{-.0in}  
    \centering
    \small
\caption{Results for the adaptation of ADEMXAPP~\cite{wu_wider_2016} from GTA to Cityscapes. The ADEMXAPP net is a more powerful semantic segmentation network than FCN-8s.}
\label{ADEMXAPPresults_GTA}
\scalebox{0.75}{
\begin{tabular}{l|c|ccccccccccccccccccc}

\hline
\multirow{2}{*}[-2em]{ Method~~~~\%} & \multirow{2}{*}[-2em]{ IoU} & \multicolumn{19}{c}{\textbf{GTA2Cityscapes} Class-wise IoU}\\

 & &  \rot{bike} & \rot{fence} & \rot{wall} & \rot{t-sign} & \rot{pole} & \rot{mbike} & \rot{t-light} & \rot{sky} & \rot{bus} & \rot{rider} & \rot{veg}  & \rot{terrain} & \rot{train} & \rot{bldg} & \rot{car} & \rot{person} & \rot{truck} & \rot{sidewalk} & \rot{road}\\
 \hline\hline
 
FCN (CC) & 26.2 & 16.2 & 10.9 & 8.8 &\textbf{ 18.5 }& 23.3 & 7.0 & 13.2 & 62.7 & 5.4 &\textbf{ 19.0 }& 65.1 & 5.8 & 2.3 & 64.8 & 63.9 & 42.2 & 9.2 & 13.8 & 45.0\\ 
\textbf{ADEMXAPP (CC)} & 30.0 & 8.3 & 10.6 & 15.5 & 16.5 & 23.3 & 11.7 &\textbf{ 21.9 }& 66.5 & 10.7 & 12.6 & 74.2 & 14.1 & 3.3 & 70.2 & 58.4 & 43.1 & 14.3 & 24.2 & 70.2\\ 
\hline
FCN (CC+SP) & 30.2 & 10.4 &\textbf{ 13.6 }& 10.3 & 14.0 & 13.9 & 18.8 & 16.5 & 73.6 & 14.1 & 9.5 & 79.2 & 12.9 & 0.0 & 74.3 & 63.5 & 33.1 & 18.9 & 27.5 & 70.5\\ 
\textbf{ADEMXAPP (CC+SP)} & 34.0 & 8.8 & 12.4 & 18.4 & 15.3 & 22.5 & 16.6 & 18.5 & 73.7 &\textbf{ 24.5 }& 10.9 & 76.7 &\textbf{ 22.9 }& 0.1 & 74.3 & 72.3 & 40.3 &\textbf{ 21.3 }& 32.9 & 83.1\\ 
\hline
FCN (CC+I) & 28.5 & 7.2 & 9.4 & 11.1 & 13.4 & 23.1 & 9.6 & 15.1 & 64.6 & 5.9 & 15.5 & 71.1 & 10.3 & 3.9 & 67.7 & 62.3 & 43.0 & 14.0 & 23.0 & 71.6\\ 
\textbf{ADEMXAPP (CC+I)} & 31.4 &\textbf{ 24.2 }& 13.3 & 19.0 & 11.2 &\textbf{ 26.2 }& 10.0 & 8.2 & 62.4 & 9.0 & 18.8 & 74.0 & 14.5 & 8.1 & 68.4 & 70.2 & 41.4 & 16.3 & 26.9 & 73.6\\ 
\hline
FCN (CC+I+SP) & 31.4 & 12.0 & 13.2 & 12.1 & 14.1 & 15.3 &\textbf{ 19.3 }& 16.8 &\textbf{ 75.5 }& 19.0 & 10.0 &\textbf{ 79.3 }& 14.5 & 0.0 & 74.9 & 62.1 & 35.7 & 20.6 & 30.0 & 72.9\\ 
\textbf{ADEMXAPP (CC+I+SP)} &\textbf{\underline{ 35.7 }}& 9.6 & 13.6 &\textbf{ 21.9 }& 14.2 & 25.0 & 15.7 & 19.2 & 72.3 & 22.1 & 18.1 & 77.7 & 19.7 &\textbf{ 14.5 }&\textbf{ 77.8 }&\textbf{ 73.3 }&\textbf{ 46.4 }& 17.4 &\textbf{ 33.8 }&\textbf{ 85.0}\\ 
\hline
\end{tabular}
}
\end{adjustwidth}
\end{table*}

\subsection{What is the ``market value'' of the synthetic data?}
Despite the positive results thus far for our curriculum domain adaptation from simulation to reality for the semantic segmentation of urban scenes, we argue that the significance of our work is in its capability to complement training sets of real data, rather than replacing them. In the long run, we expect that learning from both simulation and reality will alleviate the strong dependency of deep learning models on massive labeled real training data. Therefore, it is interesting to evaluate the ``market value'' of the synthetic data in terms of the labeling effort: how many real training images can the GTA or SYNTHIA dataset obviate in order to achieve about the same level of segmentation accuracy? 


In order to answer the above question, we design the following experiment. 
We train two versions of the VGG-19-FCN-8s network. One is trained on a portion of annotated Cityscapes images while another one is trained using the same subset of Cityscapes images plus the entire SYNTHIA dataset.  The subset is sampled from 2380 Cityscapes training images in the experiment since the remaining ones are reserved for the validation purpose. We report both models' performances under different   percentages subsampled from  Cityscapes.

Table~\ref{Mixing_training} presents the results. First of all, it is somewhat surprising to see  that even as few as five Cityscapes training images added to the SYNTHIA training set can significantly boost the results obtained from only synthetic images (from 22.0\% to 33.8\%). Second, the ``N/A'' results in the table mean that the corresponding neural networks either give rise to random predictions or have numerical issues. Note that such phenomena happen until there are more than 450 target images for the training without any synthetic images, implying that the ``market value'' of the SYNTHIA training set is at least worth 450 well-labeled real images. Actually, if we compare the results of the two rows, the network trained from the mixed training set outperforms the one from the real images only up to the 50\% mix. In other words, augmenting the Cityscapes training set with the SYNTHIA training set improves performance when the Cityscapes training set is smaller than 1000 images.



\begin{table*}
\caption{IoUs after mixing different percentages of Cityscapes images into the SYNTHIA training dataset (SYN+CS), and of models trained with different percentages of Cityscapes images without any SYNTHIA images (CS only).}
\label{Mixing_training}
\centering
\scalebox{1}{
\begin{tabular}{|l|ccccccccc|}

\hline
\# CS images & None & 5 images & 1\% & 2.5\%  & 5\% & 10\% & 20\% & 50\% & 100\%\\
\hline\hline
SYN+CS & 22.0\% & 33.8\% & 38.4\% & 41.0\% & 43.0\% & 46.5\% & 48.5\% & 53.2\% & 57.3\%\\
\hline
CS only & N/A & N/A & N/A & N/A & N/A & N/A & 38.4\%  &  52.2\% & 57.8\% \\
\hline
\end{tabular}
}
\end{table*}

\section{Review of the recent works on domain adaptation for semantic segmentation}\label{SecReview}
\colorchunk{
After our work~\cite{zhang2017curriculum } published in the IEEE International Conference on Computer Vision in 2017, there have been notably a rich line of works tackling the same problem, i.e.,  domain adaptation for the semantic segmentation of urban scenes by adapting from the synthetic imagery to real images. Some of them have reported very good results. Since our approach is ``orthogonal'' in some sense to these others, one may achieve even better results by fusing our method with these new ones. In this section, we give a comprehensive review of these new methods and also present the results of fusing ours with some of them. Most of the methods resort to adversarial training to reduce the domain discrepancy. We review such methods first, followed by the others.
}

\subsection{Adversarial training based methods}

\colorchunk{
If an adversarial classifier fails to differentiate the data instances of the source domain and the target domain, the discrepancy between the two should have been eliminated in certain sense. Many methods depend on this principle and differ  on how to incorporate it to the training of the segmentation network.

\begin{description}

\item [FCNs in the wild~\cite{hoffman_fcns_2016}.] Hoffman et al.\ generalize FCNs~\cite{long_fully_2015} from the source domain of synthetic imagery to the target domain of real images for the semantic segmentation task. They employ a pixel-level adversarial loss to enforce the network to extract domain-invariant features.

\item [CyCADA~\cite{hoffman2017cycada}.] The main idea  is to transform the synthetic images of the source domain to the style of the target domain (real images) using CycleGAN~\cite{zhu2017unpaired} before feeding the source images to the segmentation network.   CycleGAN and the segmentation network are trained simultaneously.

\item [ROAD~\cite{chen2017road}.] In the Reality Oriented Adaptation (ROAD) method~\cite{chen2017road}, two losses are proposed to align the source and the target domain. The first one is called target guided distillation, which is a loss for regression  from the segmentation network's hidden layer activation of the source domain to the image features of the target domain. Here the image features are obtained by a classifier pre-trained on ImageNet. The other loss takes care of the spatial-aware adaptation. The feature map of either a source image or a target image is partitioned into non-overlapping grids. After that, a maximum mean discrepancy loss~\cite{hoffman_fcns_2016} is introduced over each grid. 

\item [MCD~\cite{saito2017maximum}.] The Maximum Classifier Discrepancy (MCD) resembles the recently popularized generative adversarial methods~\cite{GaninICML15Unsupervised}. It learns two classifiers from the source domain and maximizes their disagreement on the target images in order to detect target examples that fall out of the support of the source domain. After that, it updates the generator to minimize the two classifiers' disagreement on the target domain. By alternating the two steps in the training, it ensures that the generator gives rise to feature representations over which the source and the target domains are well aligned.


\item [LSD~\cite{sankaranarayanan2017unsupervised}.] This work is an adversarial domain adaptation network built upon an auto-encoder network. The network takes as input both source and target images and  reconstructs them due to an auto-encoder loss. Meanwhile, an intermediate layer is connected to the segmentation network whose loss is defined using the labeled source images. 

\item [AdaptSegNet~\cite{tsai2018learning}.] Similar to FCN in the wild~\cite{hoffman_fcns_2016}, this work also employs the adversarial feature learning over the base segmentation model. Instead of having only one discriminator over the feature layer, Tsai et al.\ propose to install another discriminator on one of the intermediate layers as well. Essentially, features of different scales are forced to align.

\item [CGAN~\cite{hong2018conditional}.] This work proposes to add a fully convolutional auxiliary pathway to inject random noise into the source domain. Hence, what the segmentation network receives is the source images with perturbations. The authors found such a structure, which is motivated by the conditional GAN, greatly boosts the adaptation performance.

\item [ADR~\cite{saito2017adversarial}.] In ADR, the pixel classifier also serves as the domain classifier. It employs dropout to avoid generating features near the classification boundaries so as to avoid ambiguity to the classifier. 

\item [NMD~\cite{chen2017no}.] Besides the global adversarial feature learning module, NMD proposes a local class-wise adversarial loss over image grids. Each image grid is associated with a label distribution. The class-wise adversarial learning then tries to differentiate the source domain's label distributions from those of the target domain due to the semantic segmentation network. 

\item [DAM~\cite{huang2018domain}.] Huang et al.\ train two separate networks for the source and target domains, respectively. Since there is no segmentation annotation in the target domain, the target-domain network is trained by both regressing to the source network's weights and an adversarial loss over every layer of the two networks.

\item [FCAN~\cite{Zhang_2018_CVPR}.] Zhang et al.\ apply the adversarial loss to the lower layers of the segmentation network in addition to the common practice of using it over the last one or a few layers. The intuition is that it plays a complementary role because the lower layers mainly capture the appearance information of the images.

\item [DCAN~\cite{wu2018dcan}.] DCAN is a two-stage end-to-end network. In the first stage, it is adversarially trained to transfer the source (synthetic) images to the target (real) style. In its second stage, adversarially learning aligns the intermediate features of the two domains. Unlike the other methods, its adversarial alignment  only accounts for channel-wise features.

\item [I2I~\cite{murez2017image}.] Similar to DCAN, I2I is another domain adaptation method that learns domain agnostic features by training both an image translation network and a segmentation network.

\item [CLoss~\cite{zhu2018penalizing}.] Zhu et al.\  introduce a conservative loss in addition to the adversarial training. The conservative loss prevents overfitting the model to the source domain by penalizing overly confident source domain predictions during training.

\end{description}

\subsection{Other methods}
The adversarial training is so popular that there are only a few methods which are out of the adversarial vein for the domain adaptation of semantic segmentation. 

\begin{description}

\item [IBN~\cite{pan2018two}.] Pan et al. show that a careful balance between the instance normalization and batch normalization could enhance a neural network's cross-domain generalization. 

\item [EUSD~\cite{saleh2018effective}.] Arguing that object detectors have better generalization capacity in detecting foreground objects (e.g., car, pedestrian, etc.) than the background, Saleh et al.\ propose a simple yet powerful domain generalization segmentation framework by fusing Mask-RCNN's  detection results of the foreground~\cite{he2017mask} and DeepLab's segmentation results of the background~\cite{chen2018deeplab}.

\item [DAN~\cite{romijnders2018domain}.] As normalization (e.g., batch normalization) plays a key role in many neural semantic segmentation networks, DAN improves their performance in the target domain by simply replacing the normalization parameters with the statistics of the target domain. This change of normalization boosts the network's results in the target domain.

\item [CBST~\cite{Zou_2018_ECCV}.] Similar to our approach, this paper proposes a curriculum learning method for domain adaptation of semantic segmentation. They introduce a class-balanced self-training strategy: the most confident predictions by a model on the unlabeled instances are likely to be correct. In each round of the training process, CBST selects some of the most confident predictions on  the target images and include them in the training set of the next round.

\end{description}

\subsection{Results reported in the papers}
We summarize the results reported in the papers, along with ours, in Table~\ref{Tresults_SOTA}. Immediately, we can see that the backbone network has a huge impact on both the source-only performance without applying any adaptation techniques and the relative gain after adaptation. For instance, the performance gain of MCD jumps from 3.9\% to 17.5\% after switching the backbone network from VGG to DRN~\cite{Yu2017}. 

Another interesting observation is that the results of no adaptation vary a lot even when the same backbone network (e.g., VGG16) is used, implying that subtle changes to the implementation (e.g., removing batch normalization in order to save computation cost) can result in big differences among the final results. Through private communications with some authors of these papers, we also learned that the image resolution is another key factor. In general, higher resolution of the input image gives rise to better results no matter with or without domain adaptation. 





\subsection{The existing methods and ours are complementary}
In this section, we provide experimental evidence to show that our method is complementary to most existing methods. A late fusion of our method with another can yield better results than either individually. This is not surprising as our curriculum domain adaptation approach guides the network towards the target domain by inferring properties of the labels, whereas most existing methods learn domain-invariant features or image styles.


First of all, we analyze the class-wise prediction accuracy (evaluated by mIoU) of different methods on the target domain. Figure~\ref{fMethodComp} shows the pairwise comparison between our method and some other representative methods. The entry $(i,j)$ shows the number of classes by which the \mbox{$i$-th} method outperforms the \mbox{$j$-th} on the target domain. This table is derived from the results of  GTA2Cityscapes. Our method here is the GTA model trained without color constancy. The class-wise comparison matrix is best understood if we recall the results in Table~\ref{Tresults_SOTA}. Take I2I for instance. While it outperforms ours by 2.9\% in mIoU, ours is superior over I2I in 10 out of the 19 classes. We can draw similar observations for the other methods. To this end, we find that our approach is genuinely complementary to the other methods. Finally, we find that CBST is complementary to other adversarial training based methods as well since it is in the same vein as our curriculum domain adaptation strategy.


It is worth emphasizing that the class-wise complementary relationship between these methods is only one of the possible perspectives for the analyses. More fine-grained analyses, for example image-wise, may reveal further insights about the methods. 

Equipped with the class-wise comparison between any pair of methods, we design a simple late fusion scheme to ensemble two models. We first identify the classes for which one model gives rise to more accurate prediction than the other model by using the   validation set. Given a test image, we keep its pixel-wise labels of those classes predicted by the former model. For the remaining pixels, we label them by the latter model. We apply this late fusion scheme to CYCADA and our approach. For GTA2Cityscapes, the mIoUs of CYCADA and ours (without color constancy) are  32.5\% and 28\%, respectively. In contrast, the late fusion of the two leads to an mIoU of 34.3\% which is higher than either of them. This result and the analysis shown in Figure~\ref{fMethodComp} clearly evidence that most of the existing methods are complementary to ours for domain adaptation of the semantic segmentation task. 

}

\begin{table*}
    \centering
    \caption{\colorchunk{Comparison with the recent works published after the conference version of our approach~\cite{zhang2017curriculum}.}}\label{Tresults_SOTA}
\begin{threeparttable}

\begin{tabular}{|l||c|c|cc|cc|c|c|}

\hline
\multirow{2}{*}[-0.1em]{Method} & \multirow{2}{*}[-0.1em]{\makecell{Backbone \\ Network}} & \multirow{2}{*}[-0.1em]{\makecell{Base \xmark\\Adapt \cmark}} & \multicolumn{2}{c}{\textbf{SYN2CS}} &  \multicolumn{2}{c|}{\textbf{GTA2CS}}&\multirow{2}{*}[-0.1em]{Mechanism}&\multirow{2}{*}[-0.1em]{Code}\\
 & & & IoU & Gain & IoU & Gain & &\\
 \hline\hline

\multirow{4}{*}{CyCADA~\cite{hoffman2017cycada}} & \multirow{2}{*}{VGG-16} & \xmark & - & \multirow{2}{*}{-} & 17.9 & \multirow{2}{*}{17.5}&\multirow{4}{*}{Adversary}& \multirow{4}{*}{\href{https://github.com/jhoffman/cycada_release}{Link}}\\
& & \cmark & - &  & 35.4 &  & &\\
& \multirow{2}{*}{DRN-26} & \xmark & - & \multirow{2}{*}{-} & 21.7 & \multirow{2}{*}{17.8} & &\\
& & \cmark & - &  & 39.5 &  & &\\

\hline

\multirow{2}{*}{ROAD~\cite{chen2017road}} & \multirow{2}{*}{VGG-16} & \xmark & 25.4 & \multirow{2}{*}{10.8} & 21.9 & \multirow{2}{*}{13.0}& \multirow{2}{*}{Adversary}& \multirow{2}{*}{-}\\
& & \cmark & 36.2 &  & 35.9 & & & \\

\hline

\multirow{4}{*}{MCD~\cite{saito2017maximum}} & \multirow{2}{*}{VGG-16} & \xmark & - & \multirow{2}{*}{-} & 24.9 & \multirow{2}{*}{3.9}& \multirow{4}{*}{Adversary}& \multirow{4}{*}{\href{https://github.com/mil-tokyo/MCD_DA/tree/master/segmentation}{Link}}\\
& & \cmark & - &  & 28.8 &  & & \\
& \multirow{2}{*}{DRN-105} & \xmark & 23.4 & \multirow{2}{*}{13.9} & 22.2 & \multirow{2}{*}{17.5} & & \\
& & \cmark & 37.3 &  & 39.7 & & & \\

\hline

\multirow{2}{*}{LSD~\cite{sankaranarayanan2017unsupervised}} & \multirow{2}{*}{VGG-16} & \xmark & 26.8 & \multirow{2}{*}{9.3} & 29.6 & \multirow{2}{*}{7.5}& \multirow{2}{*}{Adversary}& \multirow{2}{*}{\href{https://github.com/swamiviv/LSD-seg}{Link}}\\
& & \cmark & 36.1 &  & 37.1 &  & & \\

\hline

\multirow{4}{*}{AdaptSegNet~\cite{tsai2018learning}} & \multirow{2}{*}{VGG-16} & \xmark & - & \multirow{2}{*}{-} & - & \multirow{2}{*}{-}& \multirow{4}{*}{Adversary}& \multirow{4}{*}{\href{https://github.com/wasidennis/AdaptSegNet}{Link}}\\
& & \cmark & 37.6\tnote{**}  &  & 35.0 &  & & \\
& \multirow{2}{*}{ResNet-101} & \xmark & 38.6\tnote{**} & \multirow{2}{*}{9.1} & 36.6 & \multirow{2}{*}{5.8}  & & \\
& & \cmark & 46.7\tnote{**}  &  & 42.4 &  & & \\

\hline

\multirow{2}{*}{FCAN~\cite{Zhang_2018_CVPR}} & \multirow{2}{*}{\makecell{ResNet-101}} & \xmark & - & \multirow{2}{*}{-} & 29.2 & \multirow{2}{*}{17.4}& \multirow{2}{*}{Adversary}& \multirow{2}{*}{-}\\
& & \cmark & - &  & 46.6 &  & & \\

\hline

\multirow{2}{*}{ADR~\cite{saito2017adversarial}} & \multirow{2}{*}{\makecell{ResNet-50}} & \xmark & - & \multirow{2}{*}{-} & 25.3 & \multirow{2}{*}{8.0}& \multirow{2}{*}{Adversary}& \multirow{2}{*}{-}\\
& & \cmark & - &  & 33.3 &  & & \\

\hline

\multirow{4}{*}{I2I~\cite{murez2017image}} & \multirow{2}{*}{ResNet-34} & \xmark & - & \multirow{2}{*}{-} & 21.1 & \multirow{2}{*}{10.7}& \multirow{4}{*}{Adversary}& \multirow{4}{*}{-}\\
& & \cmark & - &  & 31.8 &  & & \\
& \multirow{2}{*}{DenseNet-121} & \xmark & - & \multirow{2}{*}{-} & 29.0 & \multirow{2}{*}{6.7}  & & \\
& & \cmark & - &  & 35.7 &  & & \\

\hline

\multirow{6}{*}{DCAN~\cite{wu2018dcan}} & \multirow{2}{*}{VGG-16} & \xmark & 25.9 & \multirow{2}{*}{9.5} & 27.8 & \multirow{2}{*}{8.4}& \multirow{6}{*}{Adversary}& \multirow{6}{*}{-}\\
& & \cmark & 35.4 &  & 36.2 &  & & \\
& \multirow{2}{*}{\makecell{ResNet-101}} & \xmark & 28.0 & \multirow{2}{*}{8.5} & 29.8 & \multirow{2}{*}{8.7}  & & \\
& & \cmark & 36.5 &  & 38.5 &  & & \\
& \multirow{2}{*}{\makecell{PSPNet}} & \xmark & 29.5 & \multirow{2}{*}{8.9} & 33.3 & \multirow{2}{*}{8.4}  & & \\
& & \cmark & 38.4 &  & 41.7 &  & & \\

\hline

\multirow{6}{*}{DAM~\cite{huang2018domain}} & \multirow{2}{*}{VGG-16} & \xmark & 22.0\tnote{*} & \multirow{2}{*}{8.7\tnote{*}} & 18.8 & \multirow{2}{*}{13.8}& \multirow{6}{*}{Adversary}& \multirow{6}{*}{\href{https://github.com/RsEnts/DAM_release}{Link}}\\
& & \cmark & 30.7 &  & 32.6 &  & & \\
& \multirow{2}{*}{\makecell{DRN-26}} & \xmark & - & \multirow{2}{*}{-} & - & \multirow{2}{*}{-}  & & \\
& & \cmark & - &  & 40.2 &  & & \\
& \multirow{2}{*}{\makecell{ERFNet}} & \xmark & - & \multirow{2}{*}{-} & 15.8 & \multirow{2}{*}{15.5}  & & \\
& & \cmark & - &  & 31.3 &  & & \\

\hline

\multirow{2}{*}{CGAN~\cite{hong2018conditional}} & \multirow{2}{*}{VGG-16} & \xmark & 17.4 & \multirow{2}{*}{23.8} & 21.1 & \multirow{2}{*}{23.4}& \multirow{2}{*}{Adversary}& \multirow{2}{*}{-}\\
& & \cmark & 41.2 &  & 44.5 &  & & \\

\hline

\multirow{2}{*}{NMD~\cite{chen2017no}} & \multirow{2}{*}{\makecell{Dialation \\ Frontend}} & \xmark & 30.7\tnote{**} & \multirow{2}{*}{5.0} & - & \multirow{2}{*}{-}& \multirow{2}{*}{Adversary}& \multirow{2}{*}{-}\\
& & \cmark & 35.7\tnote{**} &  & - &  & & \\

\hline

\multirow{2}{*}{CLoss~\cite{zhu2018penalizing}} & \multirow{2}{*}{\makecell{VGG-16}} & \xmark & 24.9 & \multirow{2}{*}{9.3} & 30.0 & \multirow{2}{*}{8.1}& \multirow{2}{*}{Adversary}& \multirow{2}{*}{-}\\
& & \cmark & 34.2 &  & 38.1 & & &\\

\hline

\multirow{2}{*}{FCN Wld~\cite{hoffman_fcns_2016}} & \multirow{2}{*}{\makecell{Dialation \\ Frontend}} & \xmark & 17.4 & \multirow{2}{*}{2.8} & 21.1 & \multirow{2}{*}{6.0}& \multirow{2}{*}{Adversary}& \multirow{2}{*}{-}\\
& & \cmark & 20.2 &  & 27.1 & & &\\

\hline

\multirow{4}{*}{CBST~\cite{Zou_2018_ECCV}} & \multirow{2}{*}{VGG-16} & \xmark & 22.6 & \multirow{2}{*}{12.8} & 24.3 & \multirow{2}{*}{11.8}& \multirow{4}{*}{Self-Training}& \multirow{4}{*}{\href{https://github.com/yzou2/cbst}{Link}}\\
& & \cmark & 35.4 &  & 36.1 &  & & \\
& \multirow{2}{*}{ResNet-38} & \xmark & 29.2 & \multirow{2}{*}{13.3} & 35.4 & \multirow{2}{*}{11.6}  & & \\
& & \cmark & 42.5 &  & 47.0 &  & & \\

\hline

\multirow{2}{*}{IBN~\cite{pan2018two}} & \multirow{2}{*}{\makecell{ResNet-50}} & \xmark & - & \multirow{2}{*}{-} & 22.2 & \multirow{2}{*}{7.4}& \multirow{2}{*}{Normalization}& \multirow{2}{*}{\href{https://github.com/XingangPan/IBN-Net}{Link}}\\
& & \cmark & - &  & 29.6 &  & & \\

\hline

\multirow{2}{*}{EUSD~\cite{saleh2018effective}} & \multirow{2}{*}{\makecell{DeepLab\\Mask R-CNN}} & \xmark & - & \multirow{2}{*}{-} & 31.3 & \multirow{2}{*}{11.2}& \multirow{2}{*}{\makecell{Ensemble\\Detector \& Segmenter}}& \multirow{2}{*}{-}\\
& & \cmark & - &  & 42.5 &  & & \\

\hline

\multirow{2}{*}{DAN~\cite{romijnders2018domain}} & \multirow{2}{*}{\makecell{ResNet-50}} & \xmark & - & \multirow{2}{*}{-} & 34.8 & \multirow{2}{*}{3.4}& \multirow{2}{*}{Normalization}& \multirow{2}{*}{\href{https://github.com/RobRomijnders/dan}{Link}}\\
& & \cmark & - &  & 38.2 &  & & \\

\hline

\multirow{8}{*}{Ours} & \multirow{2}{*}{VGG-19} & \xmark & 22.0 & \multirow{2}{*}{7.7} & 22.3 & \multirow{2}{*}{9.1}&\multirow{8}{*}{Curriculum}& \multirow{8}{*}{\href{https://github.com/YangZhang4065/AdaptationSeg}{Link}}\\
& & \cmark & 29.7 &  & 31.4 & & &\\
& \multirow{2}{*}{VGG-19\tnote{***}} & \xmark & 22.0 & \multirow{2}{*}{7.6} & - & \multirow{2}{*}{-}  & &\\
& & \cmark & 29.6 &  & - &  & &\\
& \multirow{2}{*}{DRN-26} & \xmark & 21.9 & \multirow{2}{*}{6.3} & - & \multirow{2}{*}{-}  & &\\
& & \cmark & 28.2 &  & - &  & &\\
& \multirow{2}{*}{ADEMXAPP} & \xmark & - & \multirow{2}{*}{-} & 30.0 & \multirow{2}{*}{5.7}  & &\\
& & \cmark & - &  & 35.7 &  & &\\

\hline
\end{tabular}
\begin{tablenotes}\footnotesize
\item [*] DAM did not report the baseline performance of the SYNTHIA2Cityscapes experiment. Since their model is fine-tuned from our baseline, we report our baseline performance instead.
\item [**] NMD and AdaptSegNet used 13 SYNTHIA classes in their experiments instead of the commonly used 16.
\item [***] We do not use any external information (cf.\ Section~\ref{SecExinfor}).

\end{tablenotes}
\end{threeparttable}
\end{table*}

\begin{figure}
    \includegraphics[scale=0.45]{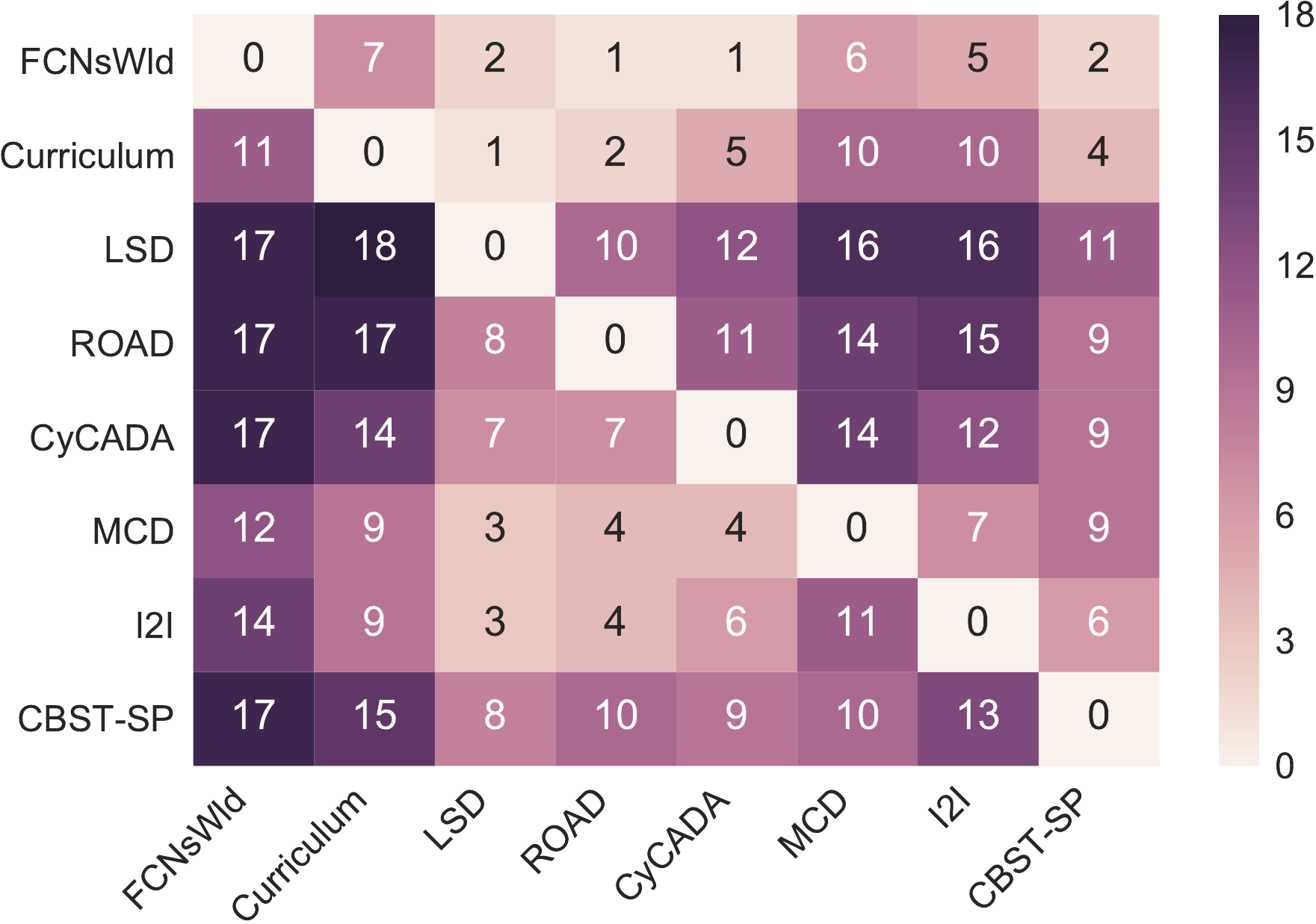}
  \caption{\colorchunk{Pairwise comparison between different domain adaptation methods for the semantic segmentation task. The entry $(i,j)$ of this table is the number of classes by which the $i$-th method outperforms the $j$-th. The results are obtained on GTA2Cityscapes. Our method is labeled \it{Curriculum}.}} \label{fMethodComp}
\end{figure}

\section{Conclusion}

In this paper, we propose a curriculum  domain adaptation approach for semantic segmentation of urban scenes. We learn to estimate the global label distributions over the target images and local label distributions over the superpixels of the target images. These tasks are easier to solve than the pixel-wise label assignment. We then use their results to effectively regularize the training of the semantic segmentation networks such that their pixel-wise predictions are consistent with the global and local label distributions. We experimentally verify the effectiveness of our approach by adapting from the source domain of synthetic images to the target domain of real images. Our method outperforms several competing baselines. Moreover, we report several key ablation studies that allow us to gain more insights about the proposed method. We also check the class-wise confusion matrices and find that some of the classes (e.g., train and bus) are almost indistinguishable in the current datasets, indicating that better simulation or more labeled real examples are required in order to achieve better segmentation results. In future work, we will explore more target properties that possess the same form as the global and local label distributions --- they are easier to solve than the pixel-wise label prediction and meanwhile can be written as a function of the pixel-wise labels. We also would like to look into the possibility of directly applying our domain adaptation framework to  virtual autonomous driving  environments such as DeepGTAV~\cite{ruano2017deepgtav } and AirSim~\cite{airsim2017fsr}.



\ifCLASSOPTIONcompsoc
  \section*{Acknowledgments}
\else
  \section*{Acknowledgment}
\fi

This work was supported by the NSF award IIS \#1566511, a gift from Adobe Systems Inc., and a GPU from NVIDIA. It was also in part supported by the NSF grant IIS-1212948, and a gift from Uber Technologies Inc.

\ifCLASSOPTIONcaptionsoff
  \newpage
\fi





\end{document}